\documentclass[lettersize,journal]{IEEEtran}

\usepackage{amsmath}
\usepackage{amsfonts}
\usepackage{amssymb}
\usepackage{mathtools}
\usepackage{bbding}
\usepackage{graphicx}
\usepackage{float}
\usepackage{newfloat}
\usepackage{wrapfig}
\usepackage[caption=false,font=footnotesize]{subfig}
\usepackage{array}
\usepackage{booktabs}
\usepackage{multirow}
\usepackage{makecell} 
\usepackage{xcolor}
\usepackage{textcomp}
\usepackage{soul}
\setuldepth{foobar}
\usepackage{listings}
\usepackage{verbatim}
\usepackage{algorithm}
\usepackage{algpseudocode} 
\usepackage{etoolbox}
\usepackage{ifthen}
\usepackage{stfloats}
\usepackage{cite}
\usepackage{url}
\newcommand{\CLA}[1]{{\color[HTML]{4472c4}\textbf{#1}}}
\newcommand{\CLB}[1]{{\color[HTML]{E76254}\textbf{#1}}}

\newcommand{\GrayCommentLine}[1]{
  \Statex\hspace{\algorithmicindent}{\color{gray!70}// #1}
}
\newcommand{\GrayCommentLineOne}[1]{
  \Statex{\color{gray!70}// #1}
}

\begin{document}

\title{Free-GVC: Toward {Retraining-Free} Extreme Generative Video Compression with Temporal Coherence}

\author{Xiaoyue Ling, Chuqin Zhou, Chunyi Li, Yunuo Chen, Yuan Tian, Guo Lu,~\IEEEmembership{Member,~IEEE}, \\ Wenjun Zhang,~\IEEEmembership{Fellow,~IEEE}
\thanks{Xiaoyue Ling, Chuqin Zhou, Yunuo Chen, Guo Lu and Wenjun Zhang are with the Institute of Image Communication and Network Engineering, Shanghai Jiao Tong University, Shanghai 200240, China (e-mail: {xiaoyue\_ling, zhouchuqin, cyril-chenyn, luguo2014, zhangwenjun}@sjtu.edu.cn). (Corresponding author: Guo Lu.)}
\thanks{Chunyi Li and Yuan Tian are with the Shanghai AI Lab, Shanghai 200232, China (e-mail: {lichunyi, tianyuan}@pjlab.org.cn)}
\thanks{\textcopyright~2026 IEEE. All rights reserved, including
rights for text and data mining and training of artificial intelligence
and similar technologies. Personal use is permitted, but
republication/redistribution requires IEEE permission.
See https://www.ieee.org/publications/rights/index.html for more information.
\par
Published in \emph{IEEE Transactions on Broadcasting}, 2026.
DOI: 10.1109/TBC.2026.3743743.}
}


\maketitle

\begin{abstract}
Building on recent advances in video generation, generative video compression has emerged as a new paradigm for achieving visually pleasing reconstructions. However, existing methods exhibit limited exploitation of temporal correlations, causing noticeable flicker and degraded temporal coherence at ultra-low bitrates. In this paper, we propose Free-GVC, a {retraining-free} generative video compression framework that reformulates video coding as latent trajectory compression guided by a video diffusion prior. Our method operates at the group-of-pictures (GOP) level, encoding video segments into a compact latent space and progressively compressing them along the diffusion trajectory. To ensure perceptually consistent reconstruction across GOPs, we introduce an Adaptive Quality Control module that {dynamically constructs an online rate–perception surrogate model to predict the optimal diffusion step for each GOP}. In addition, an Inter-GOP Alignment module establishes frame overlap and performs latent fusion between adjacent groups, thereby mitigating flicker and enhancing temporal coherence. Experiments show that Free-GVC achieves an average of 93.29\% BD-Rate reduction in DISTS over the recent neural codec DCVC-RT, and a user study further confirms its superior perceptual quality and temporal coherence at ultra-low bitrates.
\end{abstract}


\section{Introduction}
Video compression plays a vital role in managing the growing volume of visual content across streaming, storage, and communication systems\cite{zhao_TBC_2023_bg}. Conventional approaches, from traditional standards~\cite{Sullivan_2012_TCSVT_HEVC, Bross_2021_TCSVT_VVC, Zhaoyan_TBC_2024_Video} to recent neural codecs~\cite{Li_24_CVPR_DCVCFM, Jia_25_CVPR_DCVCRT, Jiang_25_CVPR_ECVC, Guo_TBC_2025_LVC}, primarily rely on transform coding to reduce spatial and temporal redundancy by mapping pixels into compact representations. While these distortion-oriented codecs excel at preserving fidelity under moderate compression, they often fail to maintain perceptual realism at low bitrates, producing blocking artifacts and over-smoothed textures. Perception-oriented methods~\cite{Li_23_MM_HVFVC, Mentzer_22_ECCV_NVCGAN, Yang_22_IJCAI_PLVC, Pan_24_TBC_JND-LIC} shift the optimization target from pixel-level distortion to perceptual similarity, improving visual quality through adversarial training and perceptual losses. However, they still suffer from blurring under extreme compression due to the lack of strong generative priors. These limitations motivate the exploration of generative methods that can synthesize missing details and restore perceptual fidelity.

\begin{figure}[t]
    \centering
    \includegraphics[width=0.95\linewidth]{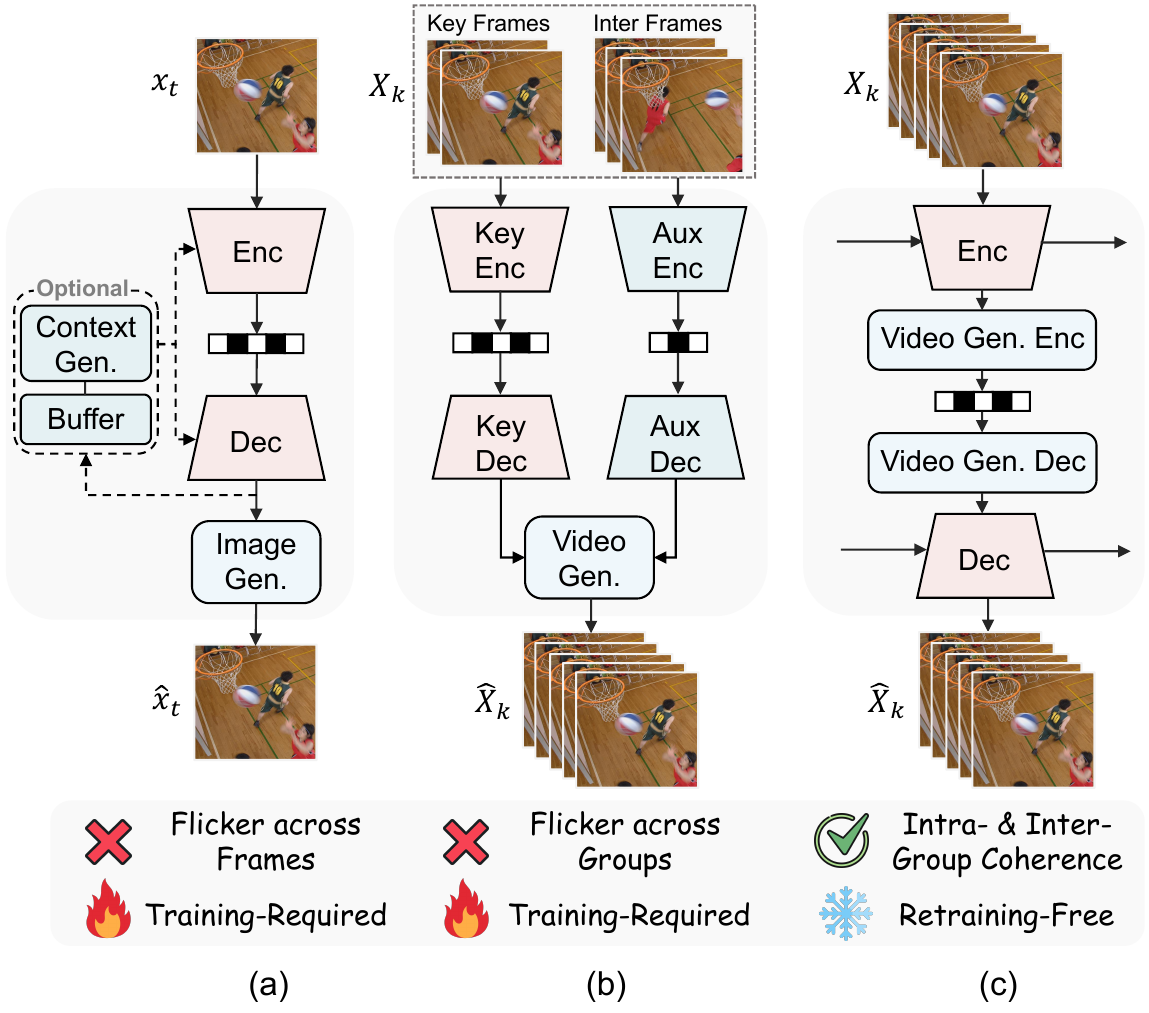}
    \caption{Comparison of generative video compression paradigms. (a) Frame-wise generation introduces frame-level flicker and requires training. (b) Group-wise generation ignores correlations across adjacent groups, leading to temporal discontinuities, and also requires training. (c) Our {retraining-free} method, {{without additional task-specific fine-tuning or retraining,}} fuses inter-group information to achieve temporally coherent compression.}
    \label{first_paradigm}
\end{figure}

Among generative approaches, diffusion models have recently emerged as a powerful paradigm for image and video synthesis~\cite{Podell_2024_ICLR_SDXL, Tim_24_Openai_Sora, Black_25_arxiv_FLUX}, owing to their remarkable ability to model complex data distributions. This capability has inspired a new line of research in generative image compression~\cite{ Li_25_TCSVT_DiffEIC, Careil_2023_ICLR_Perco, Theis_2022_arxiv_DiffC}, where pretrained diffusion priors are leveraged at the decoder to enhance highly compressed latent representations. By delegating fine-detail synthesis to these generative models, such methods achieve perceptually faithful reconstructions even at ultra-low bitrates.

\begin{figure*}[!t]
    \centering
    \includegraphics[width=1\linewidth]{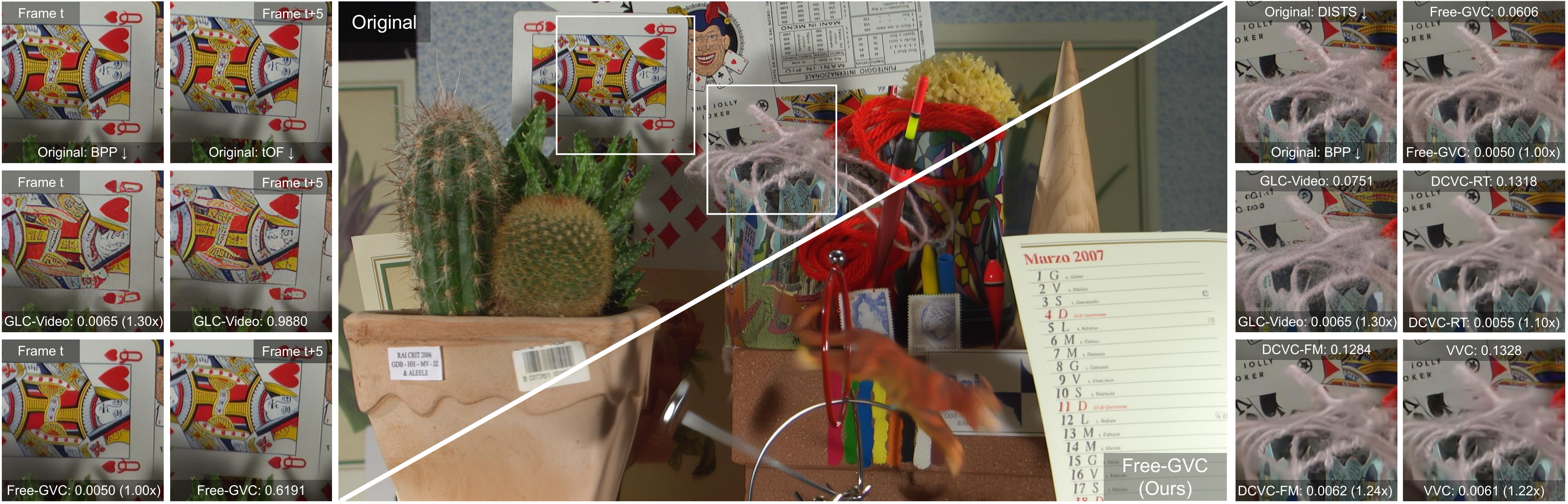}
    \caption{
    Qualitative comparison at ultra-low bitrates. The left illustrates temporal coherence measured by tOF, while the right compares perceptual quality evaluated by DISTS. Free-GVC achieves the best perceptual quality at the lowest bitrate while maintaining strong temporal consistency. Distortion-oriented codecs, including VVC~\cite{Bross_2021_TCSVT_VVC}, DCVC-RT~\cite{Jia_25_CVPR_DCVCRT}, and DCVC-FM~\cite{Li_24_CVPR_DCVCFM}, preserve temporal coherence but suffer from perceptual degradation. In contrast, generative codecs such as GLC-Video~\cite{Qi_25_TCSVT_GLCVideo} produce rich textures but exhibit misalignment with the ground truth and adjacent frames. $\downarrow$ indicates lower values are better.
    }
    \label{first}
\end{figure*}

Compared to images, videos introduce an additional temporal dimension, posing greater challenges for exploiting generative priors in compression, as illustrated in Fig.~\ref{first_paradigm}. A straightforward approach is to apply generative image codecs~\cite{Zhang_25_ICCV_StableCodec, Vonderfecht_25_ICLR_DiffC} in a frame-by-frame manner. While this yields visually rich details within each frame, it ignores inter-frame dependencies, leading to temporal flicker and inconsistency. To improve temporal coherence, some works integrate frame-wise generation into neural video compression (NVC) frameworks~\cite{Qi_25_TCSVT_GLCVideo, Ma_25_tomm_DiffVC}, where the encoder captures temporal redundancy and the decoder employs generative models for reconstruction. However, the inherent stochasticity of generative models still causes variations across frames, leaving the flicker issue unresolved. Another line of work operates at the group-of-pictures (GOP) level~\cite{Li_24_WCSP_EVCDiffusion, Wang_25_arxiv_TGVC}, improving compression efficiency by transmitting only keyframes with minimal auxiliary information, such as text prompts or optical flow, and generating the remaining frames at the decoder using video diffusion models. While this approach reduces bitrate, it often compromises the fidelity of non-keyframes. Moreover, by processing each GOP independently, these methods largely ignore correlations between neighboring segments, leading to temporal discontinuities and flicker across GOPs.

In summary, while generative approaches improve perceptual realism over conventional distortion- or perception-oriented codecs, they still struggle to maintain temporal coherence at ultra-low bitrates, as illustrated in Fig.~\ref{first}. This limitation is further exacerbated in dynamic scenes, where variations in content and motion induce inconsistent generative behavior over time. More fundamentally, diffusion-based generative compression with a fixed diffusion step is inherently ill-posed for video, as the stochastic generative process synthesizes different levels of detail across contents, preventing a consistent rate--perception trade-off and resulting in temporal quality fluctuations. Moreover, most existing methods~\cite{Zhang_25_ICCV_StableCodec, Ma_25_tomm_DiffVC, Qi_25_TCSVT_GLCVideo} require extensive retraining or fine-tuning to adapt pretrained generative models for compression.

To address these issues, we propose {Free-GVC}, a {retraining-free} generative video compression framework {that directly leverages a pre-trained video diffusion model without task-specific training or fine-tuning}. Free-GVC reformulates compression as progressive coding of noisy latent representations in the diffusion space, with the noise level corresponding to the bitrate. Specifically, the input video is divided into GOPs and projected into a compact latent space, where each group is iteratively refined along the diffusion trajectory from pure Gaussian noise to a clean reconstruction under the guidance of a pretrained video diffusion prior. This design enables compression while preserving intra-GOP temporal coherence. To ensure perceptually consistent reconstruction, we introduce an Adaptive Quality Control module that dynamically constructs an online rate--perception surrogate model during encoding to predict the appropriate diffusion step for each GOP, maintaining stable perceptual quality across diverse content. Furthermore, an Inter-GOP Alignment mechanism enhances temporal continuity by establishing frame overlap during encoding and fusing shared latent features during decoding, effectively mitigating flicker and boundary misalignment.

Our main contributions are summarized as follows:
\begin{itemize}
\item We propose {Free-GVC}, a {retraining-free} generative video compression framework that recasts video coding as diffusion-guided latent trajectory compression, enabling perceptually faithful reconstruction at ultra-low bitrates.
\item We design an Adaptive Quality Control module and an Inter-GOP Alignment mechanism, which jointly enhance perceptual stability and temporal coherence across GOPs.
\item Extensive experiments show that {Free-GVC} achieves an average of 93.29\% BD-Rate savings in DISTS across HEVC Class B$\sim$E, UVG, and MCL-JCV datasets compared with the latest neural codec DCVC-RT{\cite{Jia_25_CVPR_DCVCRT}}. {Furthermore, a user study confirms its perceptual and temporal advantages over traditional, neural, and generative baselines, even when operating at significantly lower bitrates.}
\end{itemize}

{The remainder of this paper unfolds as follows: Section~\ref{sec:related} reviews the related work on video compression and quality control. Section~\ref{sec:preliminaries} describes the background of diffusion models and the reverse channel coding (RCC) framework. Section~\ref{sec:method} provides a detailed description of our proposed Free-GVC framework, including the diffusion trajectory compression, adaptive quality control, and inter-GOP alignment mechanisms. Section~\ref{sec:results} presents extensive experimental results, comparisons, and analyses. Finally, Section~\ref{sec:conclusion} summarizes the paper and discusses potential future research directions.}

\section{Related Works}
\label{sec:related}
\subsection{Neural Video Compression}
In recent years, NVC has attracted increasing attention~\cite{Chen_25_ICCV_HyTIP, Jiang_25_CVPR_ECVC, Liao_25_MM_EHVC, Sheng_25_TIP_PRQA, Bian_25_CVPR_SEVC, Tang_25_CVPR_DCMVC, Gao_2025_ICCV_givic, Yuan_TBC_2025_LVC, Wangshen_TBC_2024_Video}. DVC~\cite{Lu_2019_CVPR_DVC} introduces the first end-to-end framework that jointly optimizes all components of the conventional hybrid coding framework. The DCVC series~\cite{Li_21_NeurIPS_DCVC, Li_2023_CVPR_DCVCDC, Li_24_CVPR_DCVCFM, Jia_25_CVPR_DCVCRT} replaces explicit residual coding with conditional mechanisms, while subsequent variants further enhance context modeling. Lu \textit{et al.}~\cite{Lu_24_AAAI_DHVC, Lu_24_arxiv_DHVC2} adopt hierarchical variational autoencoders (VAEs) to probabilistically model multi-scale latent features. Despite these advances, distortion-oriented approaches that optimize pixel-level fidelity metrics, such as MSE or PSNR, often produce overly smoothed reconstructions at low bitrates, degrading perceptual realism. To address this, several studies adopt perception-oriented training objectives. For example, DVCP~\cite{Zhang_21_VCIP_DVCP} and NVCGAN~\cite{Mentzer_22_ECCV_NVCGAN} employ LPIPS~\cite{Zhang_2018_CVPR_LPIPS} and adversarial losses~\cite{Goodfellow_2014_NeurIPS_GAN} to improve perceptual quality. PLVC~\cite{Yang_22_IJCAI_PLVC} employs a recurrent conditional discriminator, and HVFVC~\cite{Li_23_MM_HVFVC} proposes a confidence-based feature reconstruction strategy. While these methods produce more visually appealing results, the lack of strong generative priors limits their ability to preserve perceptual fidelity at ultra-low bitrates.

\subsection{Generative Video Compression}

A straightforward way to exploit generative priors for video compression is to process videos {frame-by-frame} using generative image codecs~\cite{Zhang_25_ICCV_StableCodec, Xue_25_ICCV_DLF, Jia_24_CVPR_GLC, Li_25_ICLR_GIC}. {DiffC~\cite{Vonderfecht_25_ICLR_DiffC} can apply RCC with an image diffusion prior independently to each frame, which provides strong perceptual quality for individual frames but fundamentally cannot exploit inter-frame redundancy. Since each frame is sampled independently from the image prior, fine-grained details such as textures and edges are generated without any temporal coordination, leading to flickering artifacts and temporal inconsistency. Furthermore, processing each frame with a separate RCC procedure incurs redundant coding cost for information that could otherwise be shared across frames.} To better leverage generative priors and remove temporal redundancy for video compression, a variety of dedicated generative video compression approaches have been proposed. One representative direction involves tokenizer-based methods. For instance, GLC-Video~\cite{Qi_25_TCSVT_GLCVideo} leverages VQGAN~\cite{Esser_21_CVPR_VQGAN} to extract perceptually meaningful latent features for efficient compression. Another major line of research focuses on diffusion-based approaches. DiffVC series~\cite{Ma_25_tomm_DiffVC, Ma_25_arxiv_diffvc} extend NVC architectures by integrating a VAE for feature extraction and employing an image diffusion model at the decoder to recover fine details. Other studies~\cite{Li_24_WCSP_EVCDiffusion, Wang_25_arxiv_TGVC} improve efficiency by transmitting only key frames and minimal side information, with a video diffusion model synthesizing the remaining frames at the decoder. By leveraging powerful generative priors for content reconstruction, generative compression demonstrates strong potential at ultra-low bitrates, though maintaining temporal coherence and perceptual quality remains a key challenge. Distinct from prior works requiring codec redesign or model retraining, we propose a {retraining-free} framework that directly exploits pretrained video diffusion priors for high-quality generative compression with temporally consistent reconstruction.

\subsection{Quality Control Mechanism}

Quality control has long been a fundamental challenge in video compression. Traditional codecs rely on a well-established rate–distortion optimization framework~\cite{Mao_21_TCSVT_VVCRC, Feng_24_TCSVT_RC, Liao_24_TMM_RC}, where quantization parameters are adaptively tuned to balance bitrate and reconstruction fidelity. Similar principles have been extended to NVC~\cite{Li_22_ICASSP_RC, Zhang_24_ECCV_LRC, Chen_23_TCSVT_StD}. DCVC-FM~\cite{Li_24_CVPR_DCVCFM} introduces a quantization scaler and buffer-based control mechanism to regulate frame rates. Zhang \textit{et al.}\cite{Zhang_24_ICLR_NRC} further redistribute bits across frames to enhance overall coding efficiency. Gu \textit{et al.}\cite{Gu_25_DCC_ARC} and Feng \textit{et al.}~\cite{Feng_25_TCSVT_HARC} introduce modules that predict rate–distortion curves and optimize bitrate allocation accordingly.

These approaches assume a deterministic mapping between coding parameters and reconstruction quality, an assumption that no longer holds in generative compression. The stochastic generation process produces content-dependent detail variations, leading to inconsistent perceptual quality. In this work, we explore an adaptive quality control strategy that builds an online rate–perception surrogate during encoding to select appropriate diffusion steps for each GOP, ensuring stable perceptual quality across diverse video content despite the stochastic nature of diffusion-based reconstruction.

\section{Preliminaries}
\label{sec:preliminaries}
Our compression framework builds upon reverse channel coding, a principled approach that leverages diffusion models for progressive lossy compression. This section briefly reviews its theoretical foundation and practical implementation{~\cite{Theis_22_ICML_RCC}}.

\subsection{Diffusion Models as Lossy Compression}
{Denoising Diffusion Probabilistic Model (DDPM)}~\cite{Ho_2020_NeurIPS_DDPM} can be interpreted as a form of progressive lossy source coding{\cite{Theis_2022_arxiv_DiffC}} that attempts to transmit a clean data sample $\mathbf{x}_0$, drawn from the data distribution $q(\mathbf{x}_0)$, through the forward and reverse diffusion processes. The framework relies on minimal random coding~\cite{Harsha_07_CCC_Communication, Havasi_19_ICLR_Minimal}, which enables transmitting a random variable $\mathbf{x} \sim q(\mathbf{x})$ using approximately $D_{\mathrm{KL}}(q(\mathbf{x}) \parallel p(\mathbf{x}))$ bits on average, where $p(\mathbf{x})$ is a known prior distribution for both the sender and the receiver.

{ When this principle is applied to $\mathbf{x}_0 \sim q(\mathbf{x}_0)$, the diffusion trajectory induces a progressive lossy source coding procedure. Specifically, a sequence of latent variables $\{\mathbf{x}_t\}_{t=0}^{T}$ is constructed, where $\mathbf{x}_T$ represents the fully noised latent at the final diffusion step $T$ and $\mathbf{x}_t$ represents the latent at intermediate timestep $t$. These variables are sequentially communicated from $\mathbf{x}_T$ to $\mathbf{x}_0$. At each step, the forward posterior $q(\mathbf{x}_t \mid \mathbf{x}_{t+1}, \mathbf{x}_0)$ is encoded using the learned reverse distribution $p_\theta(\mathbf{x}_t \mid \mathbf{x}_{t+1})$, where $\theta$ denotes the model parameters.}

At any intermediate timestep $t$, the receiver has full access to the partial latent $\mathbf{x}_t$ and can directly estimate the original signal $\mathbf{x}_0$ via
\begin{equation} 
\mathbf{x}_{0}\approx\hat{\mathbf{x}}_{0}=\left(\mathbf{x}_{t}-\sqrt{1-\bar{\alpha}_{t}}\boldsymbol{\epsilon}_{\theta}(\mathbf{x}_{t}, t)\right)/\sqrt{\bar{\alpha}_{t}}, 
\end{equation}
\noindent where $\boldsymbol{\epsilon}_{\theta}(\mathbf{x}_{t}, t)$ denotes the noise prediction network, $\alpha_t \coloneqq 1 - \beta_t$ is the signal preservation factor at timestep $t$, $\bar{\alpha}_t \coloneqq \prod_{s=1}^{t} \alpha_s$ is the cumulative product controlling the overall noise level, and $\beta_t \in (0,1)$ specifies a monotonically increasing variance schedule{~\cite{Ho_2020_NeurIPS_DDPM}}. Alternatively, rather than estimating $\mathbf{x}_0$ from $\mathbf{x}_t$, the receiver may stochastically reconstruct the entire reverse trajectory $\mathbf{x}_t, \mathbf{x}_{t-1}, \ldots, \mathbf{x}_0$ to achieve higher perceptual quality through the full generative process.

\begin{algorithm}[t]
\caption{\textbf{PFR} Encoding~\cite{Theis_22_ICML_RCC}}
\begin{algorithmic}[1]
\Require $p, q, w_{\min}$
\State $t, n, s^* \gets 0, 1, \infty$
\Repeat
    \State $z \gets \texttt{simulate}(n, p)$ \hfill$\triangleright$ Candidate generation
    \State $t \gets t + \texttt{expon}(n, 1)$ \hfill$\triangleright$ Poisson process
    \State $s \gets t \cdot p(z)/q(z)$ \hfill$\triangleright$ Candidate's score
    \If{$s \leq s^*$} \hfill$\triangleright$ Accept/reject candidate
        \State $s^*, n^* \gets s, n$
    \EndIf
    \State $n \gets n + 1$
\Until{$s^* \leq t \cdot w_{\min}$}
\State \Return $n^*$
\end{algorithmic}
\label{alg:PFR_encoding}
\end{algorithm}

\begin{algorithm}[t]
\caption{\textbf{PFR} Decoding~\cite{Theis_22_ICML_RCC}}
\begin{algorithmic}[1]
\Require $n^*,p$
\State \Return $\texttt{simulate}(n^*, p)$
\end{algorithmic}
\label{alg:PFR_decoding}
\end{algorithm}

\begin{figure*}[t]
    \centering
    \includegraphics[width=1\linewidth]{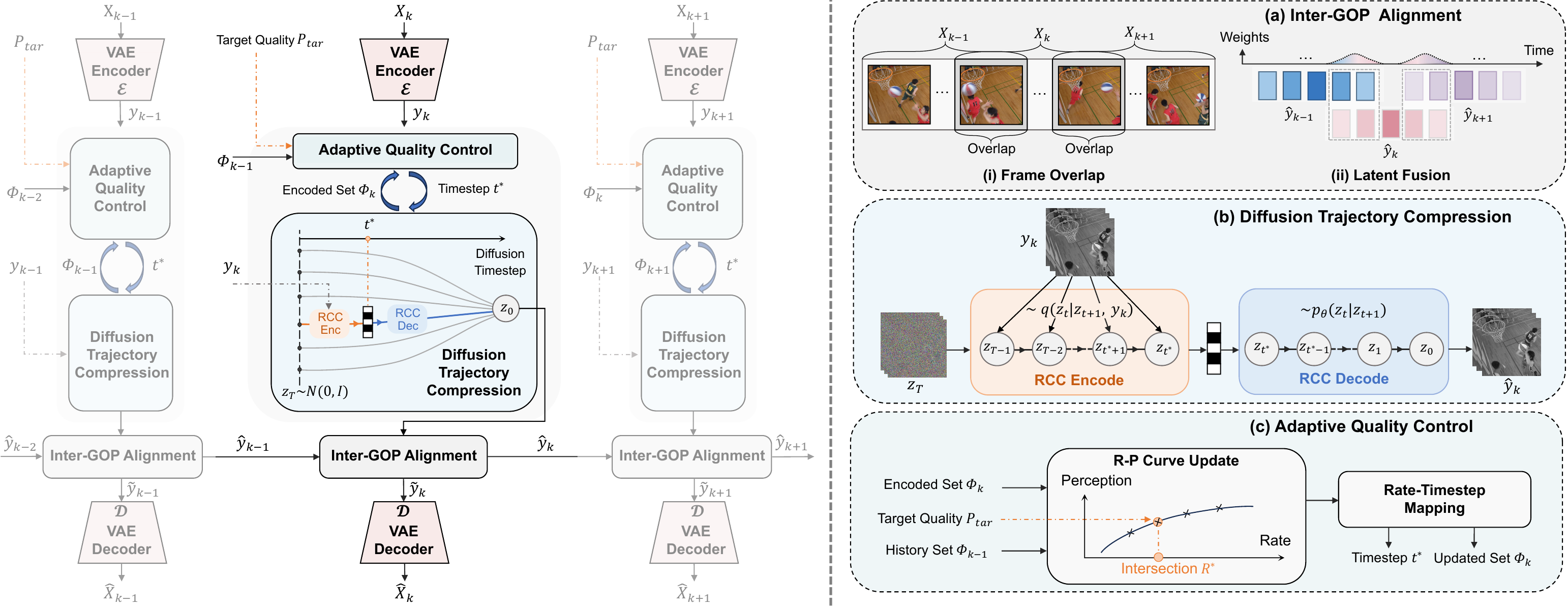}
    \caption{
    Overview of {Free-GVC} framework. (a) Inter-GOP Alignment: The input video is divided into GOPs with overlapping frames, which are fused in the latent space during decoding to ensure temporal continuity across GOP boundaries. (b) Diffusion Trajectory Compression: Each latent $\mathbf{y}_k$ is progressively coded along the diffusion trajectory using RCC to obtain $\mathbf{z}_{t^*}$, which is then denoised at the decoder. Gray curves denote possible random denoising paths, with the selected one highlighted in orange and blue. (c) Adaptive Quality Control: An online rate--perception surrogate model predicts the optimal diffusion timestep $t^*$ for each GOP to achieve the target perceptual quality.
    }
    \label{framework}
\end{figure*}

\subsection{Practical Implementation}
This lossy compression interpretation is practically instantiated through RCC~\cite{Theis_22_ICML_RCC, Theis_2022_arxiv_DiffC, Vonderfecht_25_ICLR_DiffC}, which employs the Poisson Functional Representation (PFR) algorithm~\cite{Li_18_TIT_PFR} to transmit samples from a target distribution $q$ using only a source distribution $p$, as shown in Algorithms~\ref{alg:PFR_encoding} and~\ref{alg:PFR_decoding}. PFR generates candidate samples from $p$ using different random seeds $n$ via a shared pseudorandom generator $\mathrm{simulate}$, and selects the seed $n^*$ that minimizes the likelihood-ratio score.

In our compression framework, we denote the noisy latent variables in the diffusion space as $\mathbf{z}_t$, where the target posterior distribution $q$ at timestep $t-1$ is defined as
\begin{equation}
q(\mathbf{z}_{t-1} \mid \mathbf{z}_{t}, \mathbf{y}_k)
= \mathcal{N}\!\left(\mathbf{z}_{t-1}; \tilde{\boldsymbol{\mu}}_{t}(\mathbf{z}_{t}, \mathbf{y}_k), \tilde{\beta}_{t}\mathbf{I}\right),
\end{equation}
\noindent where $\mathbf{y}_k$ represents the clean latent corresponding to the $k$-th GOP, {and $\mathbf{I}$ denotes the identity matrix}. The mean $\tilde{\boldsymbol{\mu}}_{t}(\mathbf{z}_{t}, \mathbf{y}_k)$ is given by
\begin{equation}
\tilde{\boldsymbol{\mu}}_{t}(\mathbf{z}_{t}, \mathbf{y}_k)
\coloneqq
\frac{\sqrt{\bar{\alpha}_{t-1}}\beta_t}{1-\bar{\alpha}_{t}}\, \mathbf{y}_k
+
\frac{\sqrt{\alpha_{t}}(1-\bar{\alpha}_{t-1})}{1-\bar{\alpha}_{t}}\, \mathbf{z}_{t},
\end{equation}
and the variance is
\begin{equation}
\tilde{\beta}_t \coloneqq \frac{1-\bar{\alpha}_{t-1}}{1-\bar{\alpha}_{t}}\, \beta_t.
\end{equation}

The learned reverse distribution $p_\theta$ is similarly modeled as a Gaussian distribution:
\begin{equation}
p_{\theta}(\mathbf{z}_{t-1} \mid \mathbf{z}_{t})
= \mathcal{N}\!\left(\mathbf{z}_{t-1}; \boldsymbol{\mu}_{\theta}(\mathbf{z}_{t}, t), \tilde{\beta}_{t}\mathbf{I}\right),
\end{equation}
where the mean $\boldsymbol\mu_{\theta}(\mathbf{z}_{t}, t)$ is parameterized by a neural network. Substituting the DDPM prediction $\hat{\mathbf{y}}_k = \frac{1}{\sqrt{\bar{\alpha}_{t}}}(\mathbf{z}_{t} - \sqrt{1-\bar{\alpha}_{t}}\, \boldsymbol{\epsilon}_{\theta}(\mathbf{z}_{t}, t))$ yields
\begin{align}
\boldsymbol{\mu}_{\theta}(\mathbf{z}_{t}, t)
= \frac{1}{\sqrt{\alpha_{t}}}
\left(
\mathbf{z}_{t}
- \frac{\beta_{t}}{\sqrt{1-\bar{\alpha}_{t}}}\,
\boldsymbol{\epsilon}_{\theta}(\mathbf{z}_{t}, t)
\right),
\end{align}
\noindent where $\boldsymbol{\epsilon}_{\theta}(\mathbf{z}_{t}, t)$ is used to estimate the Gaussian noise component in $\mathbf{z}_t$.

\section{Proposed Method}

\label{sec:method}

\subsection{Overview}
The overall architecture of the proposed framework, termed Free-GVC, is shown in Fig.~\ref{framework}. Given an input video sequence $\mathbf{X} \in \mathbb{R}^{L \times H \times W \times 3}$ with $L$ frames, where $H$ and $W$ denote the spatial height and width of each frame, it is first divided into $K$ GOPs $\{\mathbf{X}_1, \mathbf{X}_2, \dots, \mathbf{X}_k, \dots, \mathbf{X}_K\}$, each containing $l$ consecutive frames. To mitigate discontinuities at GOP boundaries, an Inter-GOP Alignment mechanism constructs adjacent GOPs with $m$ overlapping frames. The $k$-th GOP $\mathbf{X}_k$ is then encoded into a latent representation $\mathbf{y}_k \in \mathbb{R}^{(l/s) \times (H/d) \times (W/d) \times C}$ using the VAE encoder $\mathcal{E}$ of the pretrained diffusion model, where $s$ and $d$ denote the temporal and spatial downsampling factors, respectively, and $C$ denotes the latent channel number.

During encoding, the Adaptive Quality Control module maintains an online rate–perception model, initialized from previous GOPs and continuously updated with current observations, to predict the optimal diffusion step $t^*$ for the target quality. The clean latent $\mathbf{y}_k$ is progressively encoded along the diffusion  trajectory from timestep $T$ to $t^*$, yielding the noisy latent $\mathbf{z}_{t^*}$, which is then compressed into a bitstream using RCC~\cite{Theis_22_ICML_RCC}. At the decoder side, the transmitted latent $\mathbf{z}_{t^*}$ is denoised to obtain the clean latent $\hat{\mathbf{y}}_{k}$. The Inter-GOP Alignment mechanism then fuses latents in the overlapping regions, effectively suppressing flicker and boundary artifacts to ensure smooth transitions across GOPs. Finally, the fused latent $\tilde{\mathbf{y}}_{k}$ is passed through the VAE decoder $\mathcal{D}$ to produce perceptually faithful reconstructions $\hat{\mathbf{X}}_k$.

\begin{figure*}[t]
    \centering
    \includegraphics[width=1\linewidth]{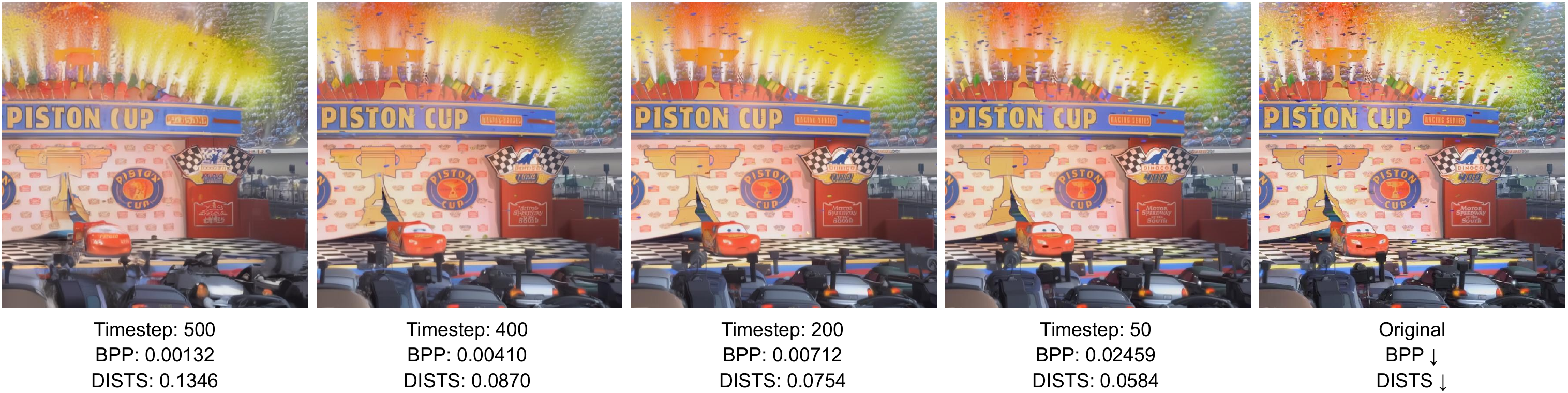}
    \caption{Progressive compression process of frame 24 from the videoSRC25 sequence in the MCL-JCV dataset{~\cite{Wang_16_ICIP_MCL-JCV}}.}
    \label{progress}
\end{figure*}

\subsection{Diffusion Trajectory Compression}\label{sec:compression}
Building on the RCC framework introduced in Section~\ref{sec:preliminaries}, we now describe how GOP $\mathbf{X}_k$ is compressed by progressively coding along the diffusion trajectory. The compression process begins from a random Gaussian sample $\mathbf{z}_T \sim \mathcal{N}(0, I)$ and proceeds toward the target latent $\mathbf{y}_k$, terminating at an adaptive timestep $t^*$ determined by the quality control module.

At each denoising step $t$ from $T$ to ${t^*}$, the encoder leverages the video diffusion prior $p_\theta$, which operates on the GOP through its spatiotemporal attention and 3D causal convolutional architecture to implicitly preserve temporal coherence. The PFR algorithm (Algorithms~\ref{alg:PFR_encoding} and~\ref{alg:PFR_decoding}) searches for optimal random seeds that align the predicted distribution $p_\theta(\mathbf{z}_t|\mathbf{z}_{t+1})$ with the true posterior $q(\mathbf{z}_t|\mathbf{z}_{t+1}, \mathbf{y}_k)$. Specifically, each candidate seed $n$ generates a latent $\mathbf{z}_n$ sampled from $p_\theta(\mathbf{z}_t|\mathbf{z}_{t+1})$ via the shared pseudorandom generator, and the encoder selects the seed $n$ whose $\mathbf{z}_n$ minimizes the divergence from the target posterior. Once the optimal seeds from step $T$ to ${t^*}$ are determined, they are entropy-coded together. {
The diffusion timestep $t^*$ is also recorded in the bitstream to specify the termination point of the progressive coding process.
}The resulting bitrate for GOP $k$ is given by:
\begin{equation}
\label{eq:rate}
R_k = \frac{\sum_{t=t^*}^{T-1} D_\mathrm{KL}(q(\mathbf{z}_t|\mathbf{z}_{t+1}, \mathbf{y}_k) \parallel p_\theta(\mathbf{z}_t|\mathbf{z}_{t+1}))}{l \times H \times W}.
\end{equation}

{
We emphasize that Eq.~\ref{eq:rate} characterizes the theoretical expected 
rate derived from the RCC formulation and does not directly represent the 
length of the actual encoded bitstream; all bitrate values reported in our experiments are instead measured from the 
actual entropy-coded files. }This approach transmits only compact random seeds rather than high-dimensional latents, significantly reducing bandwidth requirements.

At the decoder side, the transmitted seeds regenerate $\mathbf{z}_{t^*}$ using the shared diffusion prior. The partially denoised latent is then progressively refined through the remaining reverse diffusion steps ${t^*} \to 0$ to obtain the clean latent $\hat{\mathbf{y}}_{k}$. Finally, the VAE decoder reconstructs perceptually faithful frames.

An important property of this framework is its ability to provide continuous quality scaling. By adjusting the termination timestep $t^*$, the compressed bitrate can be naturally controlled. As illustrated in Fig.~\ref{progress}, at the lowest bitrate, the reconstructed frame successfully preserves the overall structure and global semantics of the scene, although fine details and subtle textures are noticeably degraded. As the bitrate increases by transmitting more diffusion steps, these discrepancies gradually diminish: finer textures, object boundaries, and high-frequency details are progressively restored, and the reconstructed frame becomes nearly indistinguishable from the original. This progressive behavior highlights the inherent capability of our diffusion-based framework to provide flexible bitrate control without retraining or additional overhead.

\begin{algorithm}[t]
\caption{\textbf{Adaptive Quality Control}}
\begin{algorithmic}[1]
\Require Latent $\mathbf{{y}_k}$, target quality $P_{tar}$, previous set $\Phi_{k-1}$
\Ensure Encoded latent $\mathbf{z}_{t^*}$, updated set $\Phi_{k}$ 

\If{$\Phi_{k-1}=\emptyset$}
    \GrayCommentLine{Sparse Timestep Sampling}
    \State Initialize $\mathbf{z}_T \sim \mathcal{N}(0, I)$
    \State Select sampling set $\mathcal{S} \subset \{1,\ldots, T\}$ with $|\mathcal{S}| = M$
    \For{$t = T-1$ to $0$}
        \State \textbf{if} $|\Phi_k| = M$ \textbf{then break}
        \State Encode $\mathbf{z}_t\sim q(\mathbf{z}_t|\mathbf{z}_{t+1},\mathbf{y}_k)$ using $p_\theta(\mathbf{z}_t|\mathbf{z}_{t+1})$
        \If{$t \in \mathcal{S}$}
            \State Compute $(R_t, P_t)$ from $\mathbf{z}_t$
            \State $\Phi_k \gets \Phi_k \cup \{(R_t, P_t)\}$
        \EndIf
    \EndFor
\Else
    \GrayCommentLine{History Information Reuse}
    \State $\Phi_k \gets \texttt{align}(\Phi_{k-1})$
\EndIf
\GrayCommentLineOne{Iterative Refinement}
\While{$|P_{t^*} - P_{tar}| > \varepsilon$} 
    \GrayCommentLine{Online Surrogate Fitting}
    \State $\alpha_k,\beta_k\gets\arg\min_{\alpha,\beta}\sum_{(R_i,P_i)\in\Phi_k} [P_i - (\alpha \cdot R_i^{\beta})]^2$
    \State $R^* \gets (P_{tar}/\alpha_k)^{\frac{1}{\beta_k}}$
    \State $t^* \gets \arg\min_t |R_t - R^*|$
    \State Encode $\mathbf{z}_{t^*}\sim q(\mathbf{z}_{t^*}|\mathbf{z}_{t^*+1},\mathbf{y}_k)$ using $p_\theta(\mathbf{z}_{t^*}|\mathbf{z}_{t^*+1})$
    \State Compute $(R_{t^*}, P_{t^*})$ from $\mathbf{z}_{t^*}$
    \State $\Phi_k \gets \Phi_k \cup \{(R_{t^*}, P_{t^*})\} $
    \State $\Phi_k \gets \Phi_k \setminus \{\arg\max_{(R_i, P_i) \in \Phi_k} |P_i - P_{tar}|\}$
\EndWhile

\State \Return $\mathbf{z}_{t^*},\Phi_{k}$
\end{algorithmic}
\label{adaptive_rd}
\end{algorithm}

{\subsection{Adaptive Quality Control}}\label{sec:control}

In diffusion-based video compression, bitrate control presents a fundamentally different challenge compared to conventional codecs. Due to content variation across GOPs and the inherent randomness of diffusion sampling, a fixed diffusion step may yield substantially different bitrates and perceptual quality for different video segments. This stochastic behavior can cause noticeable quality fluctuations across GOPs, which is particularly detrimental to temporal coherence.

To address this issue, we formulate quality control in diffusion-based compression as an online inverse rate–perception mapping problem. Given a target perceptual quality $P_{tar}$, the goal is to determine an appropriate diffusion timestep such that the expected coding rate and reconstruction quality match the target with minimal deviation. As detailed in Section~\ref{sec:compression}, our progressive encoding paradigm naturally supports rollback to any previous timestep without re-encoding, enabling efficient quality control with minimal overhead. The complete procedure is outlined in Algorithm~\ref{adaptive_rd}.

\subsubsection{Sparse Timestep Sampling}
During encoding, $M$ candidate timesteps are uniformly sampled from the diffusion schedule of $T$ steps, forming the set $S$. For each sampled timestep, the state is decoded, with the resulting bitrate $R_t$ and perceptual quality $P_t$ calculated and recorded in $\Phi_k$. This sparse sampling process provides a set of anchor points that characterize the rate–perception behavior of the current GOP, enabling efficient construction of a surrogate model without exhaustively traversing the entire diffusion trajectory.

\subsubsection{Online Surrogate Fitting}
Given the sampled rate–perception pairs $\Phi_k=\{(R_i,P_i)\}_{i=1}^M$, we construct a lightweight surrogate model to approximate the local relationship between bitrate $R$ and perceptual quality $P$ for the current GOP. Rather than assuming a global analytical form, the surrogate is fitted online and updated dynamically to adapt to content variations.

\begin{figure}[t]
  \centering
  \subfloat[BQSquare Sequence.\label{BQSquare_curve}]{
    \includegraphics[width=0.48\linewidth]{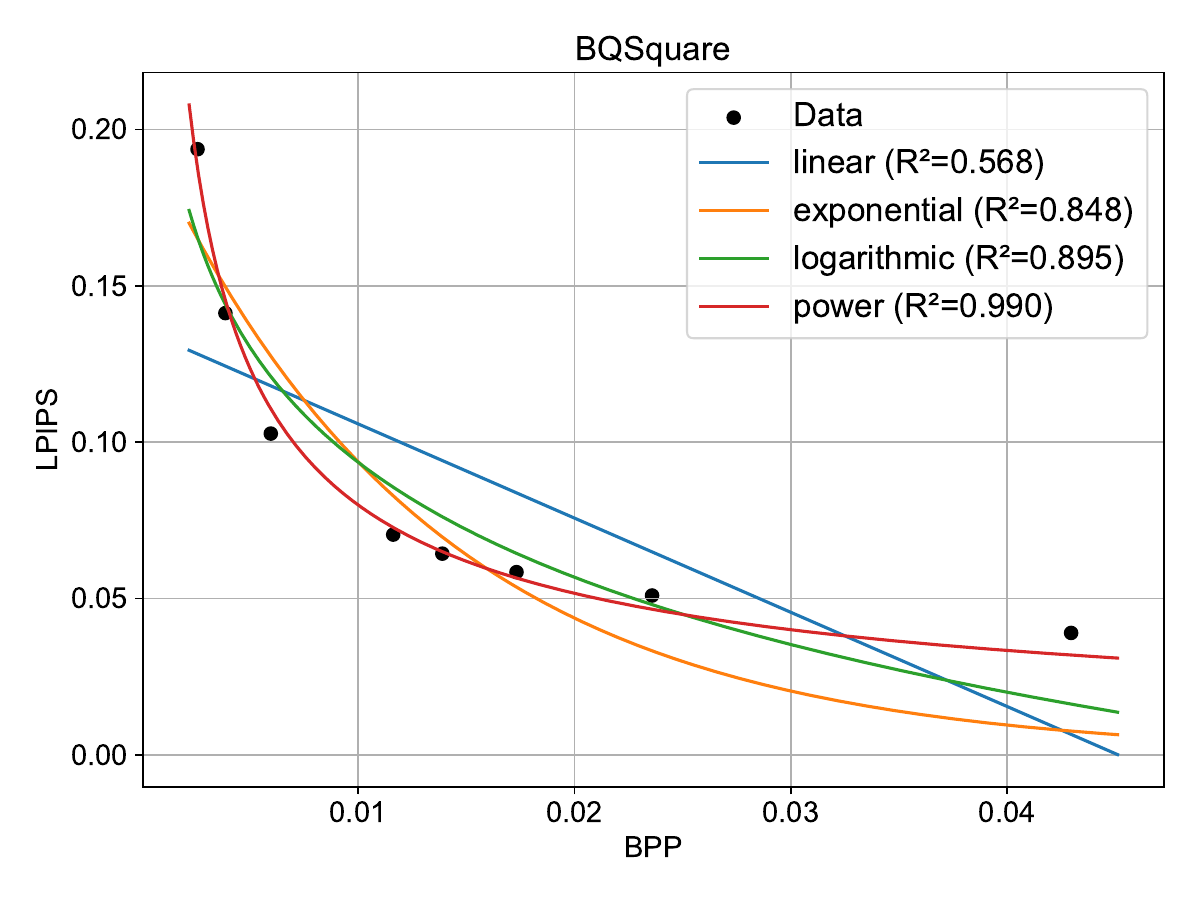}}
  \hfill
  \subfloat[BlowingBubbles Sequence.\label{BlowingBubbles_curve}]{
    \includegraphics[width=0.48\linewidth]{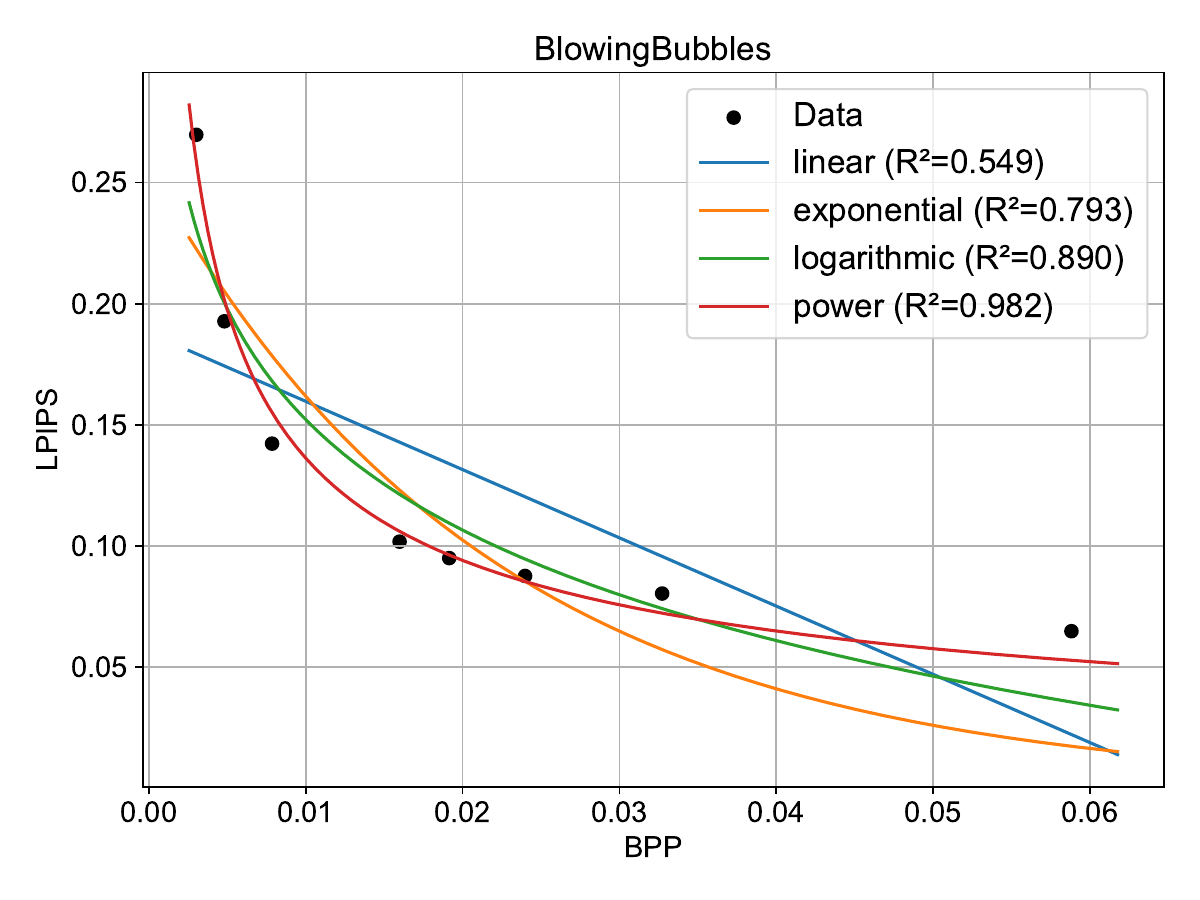}}
  \caption{Fitting performance of different models for the relationship between bitrate and perception across different sequences.}
  \label{curve-fit}
\end{figure}

We adopt a power-law formulation $P = \alpha \cdot R^{\beta}$ as the surrogate function due to its strong empirical fitting accuracy and its consistency with classical rate–distortion behavior observed in both traditional and learned codecs~\cite{Feng_24_TCSVT_RC, Liao_24_TMM_RC}. As shown in Fig.~\ref{curve-fit}, {where Fig.~\ref{curve-fit}{(a)} corresponds to a relatively static scene and Fig.~\ref{curve-fit}(b) depicts a highly dynamic scenario with frequent scene transitions,} the power model achieves consistently higher goodness-of-fit ($R^2$) across diverse video content compared to linear, logarithmic, and exponential alternatives. {The fitted model is used as an efficient inverse mapping that enables reliable prediction of diffusion step for bitrate targeting.}

The parameters $\alpha_k$ and $\beta_k$ are estimated by solving the following regression problem:
\begin{equation}
\min_{\alpha,\beta}\sum_{(R_i,P_i)\in\Phi_k} [P_i - (\alpha \cdot R_i^{\beta})]^2,
\end{equation}
\noindent where $\alpha$ and $\beta$ control the scale and nonlinearity of the fitted rate–perception curve, respectively. The corresponding target bitrate is predicted as $R^* = \left(P_\mathrm{tar}/{\alpha_k}\right)^{\frac{1}{\beta_k}}$.

\subsubsection{Iterative Refinement}

{
During RCC encoding, the latent is progressively encoded along the reverse diffusion trajectory from timestep $T$ toward $0$. Since the bitstream is generated incrementally, the accumulated coding bitrate can be measured exactly after each timestep without estimation. Given a target bitrate constraint $R^*$, the encoder continuously monitors the accumulated bitrate and terminates the encoding process once the closest achievable bitrate is reached. Formally, the selected diffusion timestep is determined by $t^*=\arg\min_t |R(t)-R^*|$, where $R(t)$ denotes the accumulated bitrate after encoding to timestep $t$. 

}

Once the target timestep $t^*$ is determined, the selected state is then decoded and evaluated, yielding the pair $(R_{t^*}, P_{t^*})$, which is added to the current observation set $\Phi_k$. This iterative refinement process can be interpreted as a feedback control mechanism, where the deviation between the achieved and target perceptual quality is progressively minimized. If the predicted perceptual quality $P_{t^*}$ deviates from the target $P_{tar}$ by more than a threshold $\varepsilon$, the model fitting and prediction steps are repeated with the updated $\Phi_k$. By continuously updating the surrogate model with newly observed samples, the control loop converges rapidly to a stable diffusion step that satisfies the target quality constraint with minimal overhead. To maintain a stable fitting model, the sample farthest from $P_{tar}$ is removed before refitting, ensuring that the most representative points guide subsequent predictions. Through empirical evaluation, we set $\varepsilon= 0.005$, which achieves stable convergence with moderate iteration complexity while maintaining consistent perceptual control across diverse content.

\subsubsection{History Information Reuse}
To accelerate convergence and reduce redundant sampling, we exploit rate–perception similarity between consecutive GOPs. Instead of directly reusing the historical set $\Phi_{k-1}$, an alignment point is first obtained by sampling a timestep $t$ in the current GOP and measuring its $(R_t, P_t)$ state. This alignment point scales and shifts $\Phi_{k-1}$ to match current GOP characteristics. The adjusted historical points are then combined with new samples to fit the updated model for GOP $k$. This history reuse strategy enables the control mechanism to exploit temporal consistency across neighboring GOPs, significantly reducing sampling overhead while preserving robustness to content changes.

\begin{figure*}[t]
    \centering
    \includegraphics[width=1\linewidth]{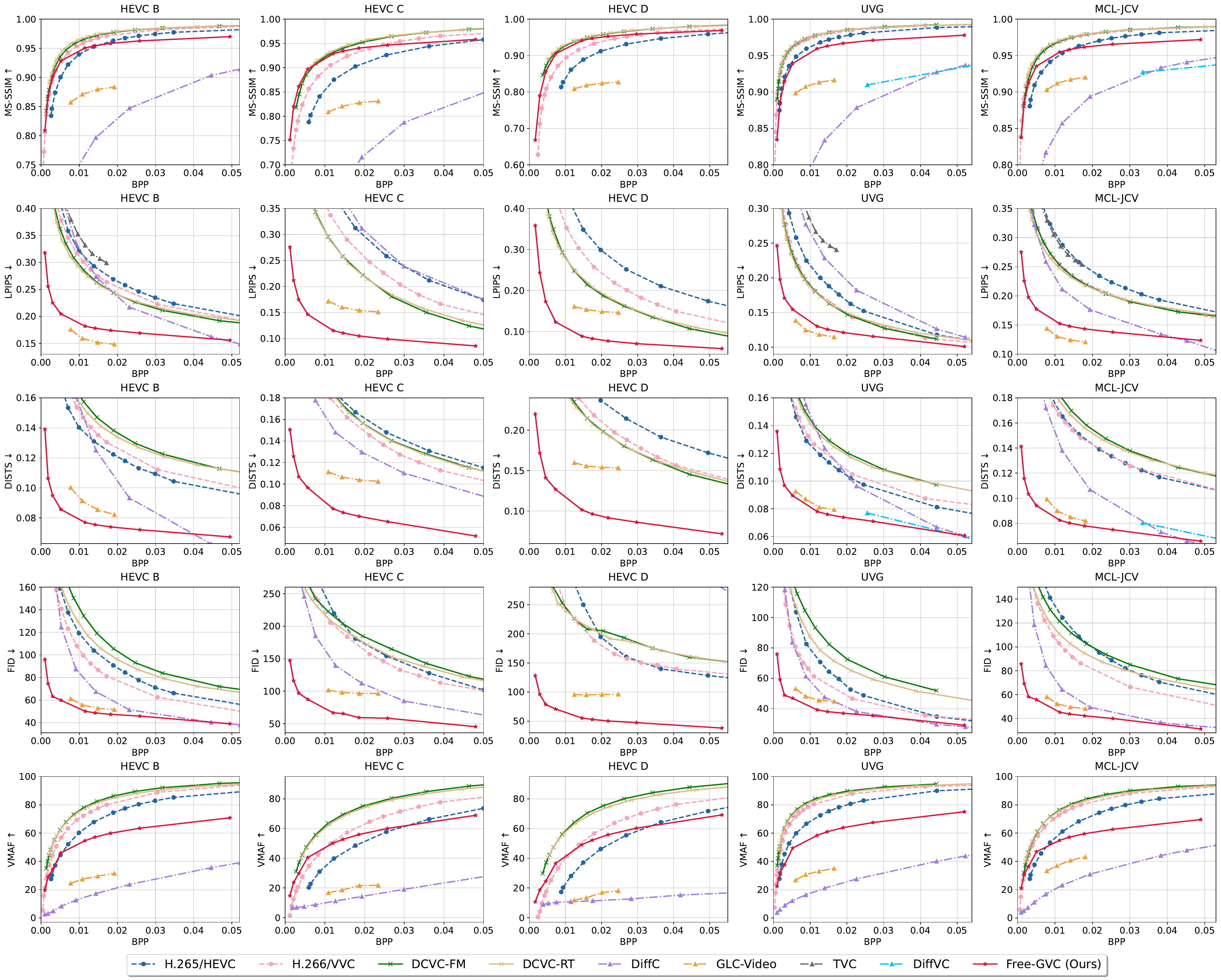}
    \caption{Rate--metric curves on the HEVC Class B--D, UVG, and MCL-JCV datasets {for H.265/HEVC~\cite{Sullivan_2012_TCSVT_HEVC}, H.266/VVC~\cite{Bross_2021_TCSVT_VVC}, DCVC-FM~\cite{Li_24_CVPR_DCVCFM}, DCVC-RT~\cite{Jia_25_CVPR_DCVCRT}, DiffC~\cite{Vonderfecht_25_ICLR_DiffC}, GLC-Video~\cite{Qi_25_TCSVT_GLCVideo}, TVC~\cite{Zhou_25_VI_tvc}, DiffVC~\cite{Ma_25_tomm_DiffVC}, and the proposed Free-GVC}.
}
    \label{main}
\end{figure*}

\subsection{Inter-GOP Alignment}

Processing each GOP independently often causes visible discontinuities at GOP boundaries, especially in dynamic video scenes. To alleviate this, we design an Inter-GOP Alignment mechanism operating jointly across the encoding and decoding stages. It consists of two complementary parts: Frame Overlap at the encoder and Latent Fusion at the decoder.

\subsubsection{Frame Overlap}
During encoding, adjacent GOPs share $m$ overlapping frames to maintain temporal continuity across boundaries. Specifically, for two consecutive GOPs $\mathbf X_{k-1}$ and $\mathbf X_k$, the last $m$ frames of $\mathbf X_{k-1}$ are reused as the first $m$ frames of $\mathbf X_k$. Each GOP is then encoded into the latent space by the video VAE encoder, which downsamples the temporal dimension by a factor of $s$. Consequently, the overlapping region in the latent domain can be denoted as $\{\hat{y}_{k-1}^{(l'-i)}, \hat{y}_{k}^{(i)}\}_{i=1}^{m'}$, where $l' = l/s$ and $m' = m/s$.

\subsubsection{Latent Fusion}
At the decoding stage, overlapping latents are blended in the latent space to ensure smooth transitions at GOP boundaries. For each overlapping index $i \in \{1, \ldots, m'\}$, we apply a weighted combination:
\begin{equation}
\tilde{y}_{k}^{(i)} = \gamma(i) \hat{y}_{k}^{(i)} + \big(1-\gamma(i)\big) \hat{y}_{k-1}^{(l'-i)},
\end{equation}
where $\gamma(i)\in[0,1]$ is the fusion weight function that gradually shifts the contribution from the previous GOP to the current one. The fused latent $\tilde{\mathbf{y}}_{k}$ is then decoded to reconstruct $\hat{\mathbf{X}}_k = \mathcal{D}(\tilde{\mathbf{y}}_{k})$, thereby effectively suppressing boundary artifacts and maintaining temporal coherence across GOPs.

Since overlapping frames are redundantly encoded, this strategy introduces a slight bitrate overhead. For a video with $K$ GOPs, each containing $l$ frames, $l \times K$ frames are encoded in total, of which $m \times (K - 1)$ are redundantly processed due to overlap. The effective bitrate $R_f$ is given by
\begin{equation}
R_f = R \cdot \frac{l \times K}{l + (K-1)(l - m)},
\end{equation}
where $R$ denotes the original bitrate. In practice, since $m \ll l$, the overhead is negligible. For $l = 48$, $m = 4$, and $K = 2$, the bitrate increases by less than 4.4\% while notably improving temporal smoothness across GOP boundaries.

It is worth noting that our coding process remains strictly GOP-wise in terms of RCC sampling and entropy coding, with no cross-GOP dependency introduced in the bitstream. {This GOP-wise design enables each GOP to be processed independently during the encoding and entropy coding stages, while Inter-GOP Alignment at the reconstruction stage introduces a lightweight dependency through latent fusion to improve temporal coherence.
}

\begin{table*}[t]
    \caption{BD-Rate (\%) and BD-Metric comparison on HEVC B--E, UVG, and MCL-JCV datasets, using DCVC-RT~\cite{Jia_25_CVPR_DCVCRT} as the anchor. $\downarrow$ and $\uparrow$ indicate that lower and higher BD-Metric values are better, respectively. Negative BD-Rate values indicate bitrate savings. Best and second-best results are highlighted in \CLB{red} and \CLA{blue}, respectively. ``N/A'' indicates that BD-Rate cannot be computed due to insufficient overlap between the rate--metric curves.}
    \centering
    \renewcommand{\arraystretch}{1.05}
    \renewcommand{\tabcolsep}{6pt}
    
    \label{tab:main}
    \resizebox{1.0\linewidth}{!}{
    \begin{tabular}{c|c|cc|cc|cc}
    \toprule[1pt]
    \multirow{2}{*}{\textbf{Dataset}} & 
    \multirow{2}{*}{\textbf{Metric}} & 
    \multicolumn{2}{c|}{\textbf{\raisebox{0.25ex}{Traditional Codec}}} & 
    \multicolumn{2}{c|}{\textbf{\raisebox{0.25ex}{Neural Codec}}} & 
    \multicolumn{2}{c}{\textbf{\raisebox{0.25ex}{Generative Codec}}}\\
    \cline{3-8}
     & & \textbf{\raisebox{-0.4ex}{HEVC~\cite{Sullivan_2012_TCSVT_HEVC}}} & \textbf{\raisebox{-0.4ex}{VVC~\cite{Bross_2021_TCSVT_VVC}}} & \textbf{\raisebox{-0.4ex}{DCVC-FM~\cite{Li_24_CVPR_DCVCFM}}} & \textbf{\raisebox{-0.4ex}{DCVC-RT~\cite{Jia_25_CVPR_DCVCRT}}} & \textbf{\raisebox{-0.4ex}{GLC-Video~\cite{Qi_25_TCSVT_GLCVideo}}} & \textbf{\raisebox{-0.4ex}{Free-GVC (Ours)}} \\
    \midrule[0.25pt]

    \multirow{4}{*}{HEVC B} 
        & MS-SSIM $\uparrow$  &  102.24 / -0.0211   & 32.74 / -0.0062   & \CLA{8.24  / -0.0035}  & \CLB{0.00 / 0.0000} & {N/A / -0.0963} & {50.47 / -0.0166} \\
        & LPIPS $\downarrow$  &  50.85 / 0.0416   & 23.76 / 0.0196   & 5.18  / 0.0046  & {0.00 / 0.0000} & \CLB{-91.74 / -0.1167} & \CLA{-87.19 / -0.1107} \\
        & DISTS $\downarrow$  & -31.68 / -0.0149 & -14.67 / -0.0068  & 12.02 / 0.0050  & 0.00 / 0.0000 & \CLA{-89.81 / -0.0603} & \CLB{-95.32 / -0.0818} \\
        &  FID $\downarrow$   &  -9.47 / -5.134 & {-33.59 / -19.326} & 28.80 / 12.533 & 0.00 / 0.0000 & \CLA{-89.22 / -62.8868} & \CLB{ -95.37 / -77.8182} \\
    \midrule[0.25pt]

    \multirow{3}{*}{HEVC C} 
        & MS-SSIM $\uparrow$  &  157.87 / -0.0488   & 61.53 / -0.0227   & \CLA{5.50  / -0.0025}  & \CLB{0.00 / 0.0000} & {N/A / -0.1245} & {31.90 / -0.0027} \\
        & LPIPS $\downarrow$ & {85.16 / 0.0841} & 36.66 / 0.0441 & {-0.10 / -0.0001} & {0.00 / 0.0000} & \CLA{-57.07 / -0.0856} & \CLB{-89.49 / -0.1567} \\
        & DISTS $\downarrow$ & 12.44 / 0.0061 & {-11.68 / -0.0088} & 2.63 / 0.0015 & 0.00 / 0.0000 & \CLA{-73.74 / -0.0610} & \CLB{-95.26 / -0.1012} \\
        &  FID $\downarrow$ & 13.64 / 7.342 & {-6.24 / -4.339} & 14.85 / 9.025 & 0.00 / 0.000 & \CLA{N/A /  -88.9025} & \CLB{-95.69 / -130.8080} \\
    \midrule[0.25pt]
    
    \multirow{3}{*}{HEVC D} 
        & MS-SSIM $\uparrow$  &  161.48 / -0.0444   & 69.74 / -0.0239   & \CLA{3.30  / -0.0016}  & \CLB{0.00 / 0.0000} & {N/A / -0.1339} & {44.82/ -0.0061} \\
        & LPIPS $\downarrow$  &  120.77 / 0.1027   & 49.26 / 0.0581   & {-0.71  / -0.0008}  & {0.00 / 0.0000} & \CLA{ -43.93/ -0.0555} & \CLB{-78.64/ -0.1334} \\
        & DISTS $\downarrow$  & 69.11 / 0.0357 & 15.73 / 0.0101  & 1.18 / 0.0009  & 0.00 / 0.0000 & \CLA{-66.58 / -0.0534} & \CLB{-89.25 / -0.1143} \\
        &  FID $\downarrow$   & 11.55 / 1.9473 & {-14.56 / -8.6531} & 4.30 / 2.3789 & 0.00 / 0.000 & \CLA{N/A / -110.9691} & \CLB{ -97.38 / -161.3781} \\
    \midrule[0.25pt]

    \multirow{3}{*}{HEVC E} 
        & MS-SSIM $\uparrow$  &  91.26 / -0.0107   & 59.17 / -0.0059   & \CLA{-9.63  / 0.0006}  & \CLB{0.00 / 0.0000} & {N/A / -0.0541} & {357.43/ -0.0133} \\
        & LPIPS $\downarrow$ & 30.49 / 0.0206 & 32.96 / 0.0189 & {-11.41 / -0.0059} & {0.00 / 0.0000} & \CLA{-80.80 / -0.0444} & \CLB{-89.55 / -0.0689} \\
        & DISTS $\downarrow$ & -9.44 / -0.0052 & {-12.59 / -0.0054} & -4.98 / -0.0022 & 0.00 / 0.0000 & \CLA{-54.36 / -0.0201} & \CLB{-89.46 / -0.0505} \\
        &  FID $\downarrow$ & 38.93 / 17.9680 & {23.93 / 10.6847} & {2.64 / 1.1222} & 0.00 / 0.000 & \CLA{-49.11 / -12.3345} & \CLB{-81.94 / -35.0579} \\
    \midrule[0.25pt]
    
    \multirow{3}{*}{UVG} 
        & MS-SSIM $\uparrow$  &  92.15 / -0.0175   & 7.55 / -0.0018   & \CLA{2.34  / -0.0009}  & \CLB{0.00 / 0.0000} & {N/A / -0.0667} & {158.26/ -0.0245} \\
        & LPIPS $\downarrow$ & 57.14/ 0.0378 & 7.43 / 0.0060 & 1.02 / 0.0009 & {0.00 / 0.0000} & \CLB{-74.69 / -0.0634} & \CLA{ -72.62 / -0.0615} \\
        & DISTS $\downarrow$ & {-26.55 / -0.0141} & -24.36 / -0.0120 & 6.25 / 0.0029 & 0.00 / 0.0000 & \CLA{-88.74 / -0.0573} & \CLB{-93.66 / -0.0708} \\
        &  FID $\downarrow$ & -7.26 / -2.371 & {-42.51 / -21.498} & 26.22 / 9.348 & 0.00 / 0.000 & \CLA{-79.02 / -40.6620} & \CLB{-93.51 / -56.1275} \\
    \midrule[0.25pt]
    
    \multirow{3}{*}{MCL-JCV} 
        & MS-SSIM $\uparrow$  &  105.05 / -0.0218   & \CLA{5.78 / -0.0018}   & {8.25  / -0.0027}  & \CLB{0.00 / 0.0000} & {N/A / -0.0583} & {63.72/ -0.0156} \\
        & LPIPS $\downarrow$ &  85.07 / 0.0538 & 10.49 / 0.0080 & 10.49 / 0.0076 & {0.00 / 0.0000} & \CLB{-91.91 / -0.1148} & \CLA{-88.28 / -0.1082} \\
        & DISTS $\downarrow$  &  -16.27 / -0.0088 & -24.88 / -0.0137 & 12.24 / 0.0066 & {0.00 / 0.0000} & \CLA{-93.51 / -0.0863} & \CLB{-96.81 / -0.1026} \\
        &  FID $\downarrow$ & 47.96 / 17.953 & {-15.20 / -6.948} & 30.70 / 10.323 & 0.00 / 0.000 & \CLA{-92.10 / -57.2647} & \CLB{-96.35 / -70.0796} \\
    
    \bottomrule[1pt]
    \end{tabular}
    }
\end{table*}

\begin{figure*}[t]
    \centering
    \includegraphics[width=1\linewidth]{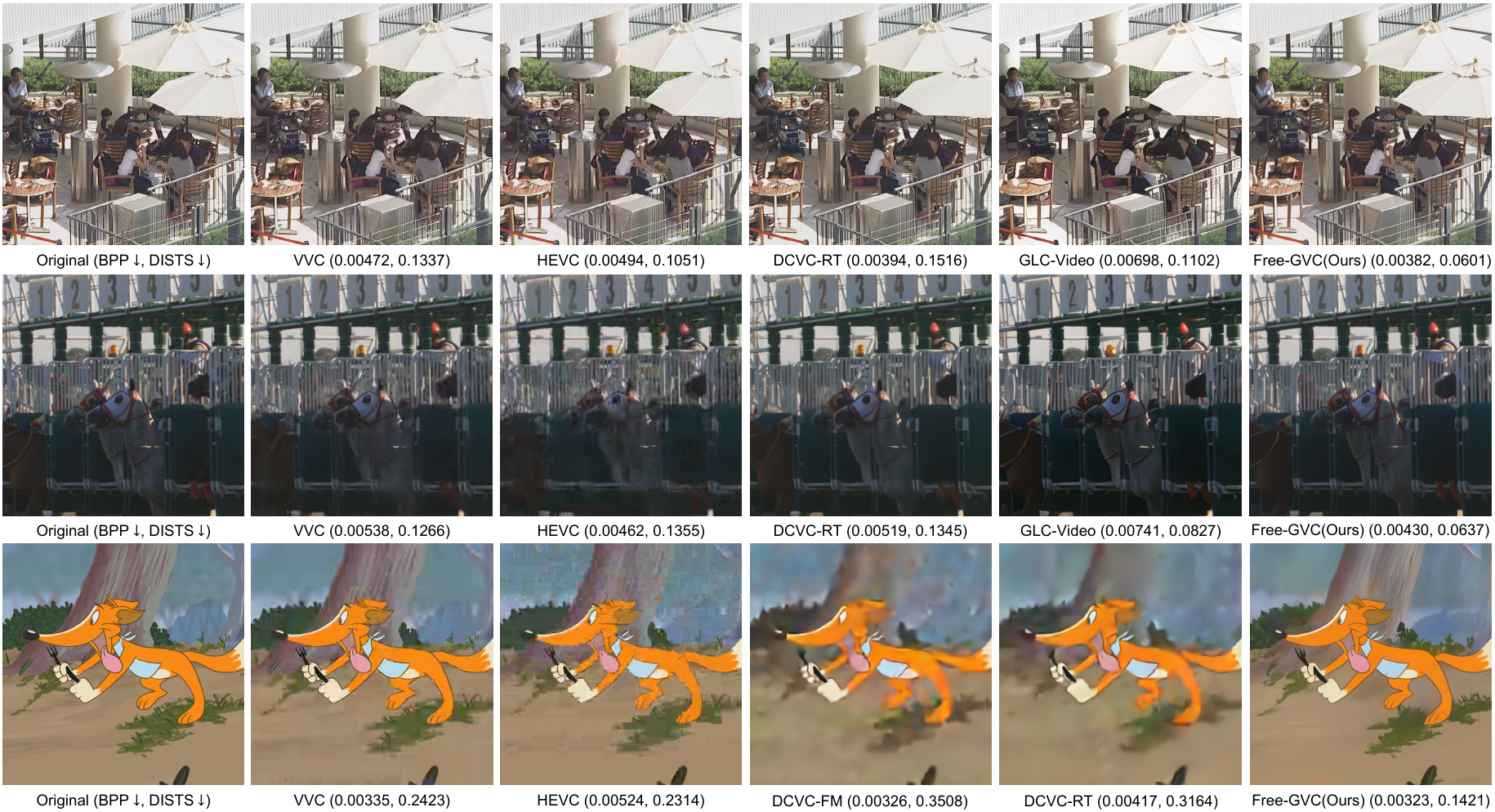}
    \caption{Visual comparison of reconstructions by {HEVC~\cite{Sullivan_2012_TCSVT_HEVC}, VVC~\cite{Bross_2021_TCSVT_VVC}, DCVC-RT~\cite{Jia_25_CVPR_DCVCRT}, DCVC-FM~\cite{Li_24_CVPR_DCVCFM}, GLC-Video~\cite{Qi_25_TCSVT_GLCVideo}, and the proposed Free-GVC}. From top to bottom: BQTerrace from HEVC Class B~\cite{Flynn_16_ITU_HEVC}, ReadySteadyGo from UVG~\cite{Mercat_20_MMsys_UVG}, and videoSRC20 from MCL-JCV~\cite{Wang_16_ICIP_MCL-JCV}. Zoom in for better comparison.}
    \label{visual}
\end{figure*}

\section{Experiments}
\label{sec:results}
\subsection{Experimental Settings}

\textbf{Evaluation Datasets.}  
We evaluate our framework on several widely used video compression benchmarks, including HEVC Class B$\sim$E\cite{Flynn_16_ITU_HEVC}, UVG\cite{Mercat_20_MMsys_UVG}, and MCL-JCV\cite{Wang_16_ICIP_MCL-JCV}.

\textbf{Comparison Methods.}  
We compare our method with three baseline categories, including traditional codecs HEVC~\cite{Sullivan_2012_TCSVT_HEVC} and VVC~\cite{Bross_2021_TCSVT_VVC}, neural codecs DCVC-FM~\cite{Li_24_CVPR_DCVCFM} and DCVC-RT~\cite{Jia_25_CVPR_DCVCRT}, and generative codecs DiffC~\cite{Vonderfecht_25_ICLR_DiffC}, {TVC~\cite{Zhou_25_VI_tvc}, DiffVC~\cite{Ma_25_tomm_DiffVC}}, and GLC-Video~\cite{Qi_25_TCSVT_GLCVideo}. 

\textbf{Evaluation Metrics.}  
Following~\cite{Li_2023_CVPR_DCVCDC, Li_24_CVPR_DCVCFM, Jiang_25_CVPR_ECVC, Qi_25_TCSVT_GLCVideo}, we evaluate the first 96 frames of each video sequence. Bit per pixel (BPP) is used to quantify the average number of bits required to encode a single pixel. To assess perceptual quality, we report LPIPS~\cite{Zhang_2018_CVPR_LPIPS} (with AlexNet features by default), DISTS~\cite{Ding_22_TPAMI_DISTS}, and FID~\cite{Heusel_2017_NeurIPS_FID}. Following~\cite{Jia_24_CVPR_GLC, Zhang_25_ICCV_StableCodec}, FID is computed on $256 \times 256$ image patches. For the HEVC Class D dataset, whose resolution ($416 \times 240$) is relatively low, we instead compute FID using $64 \times 64$ patches as in~\cite{Jia_25_CoD}. Reconstruction fidelity is evaluated using MS-SSIM~\cite{Wang_03_ACSSC_MSSSIM}, providing a distortion-oriented reference for more comprehensive comparison. {We additionally report VMAF~\cite{Li_16_Netflix_VMAF}, a full-reference video quality metric calibrated toward subjective quality assessment, to complement the structural fidelity measured by MS-SSIM.} Temporal consistency is evaluated using tOF and tLP~\cite{Oh_22_ECCV_DeMFI, Youk_2024_CVPR_FMANet}, where tOF is based on DIS estimator~\cite{Kroeger_16_ECCV_DIS} and tLP measures temporal perceptual variation via LPIPS. We also report BD-Rate and BD-Metric~\cite{Bjontegaard_01_ITU_BDRate} for overall performance comparison, where higher BD-Metric indicates better MS-SSIM and lower values are preferred for perceptual metrics. Negative BD-Rate values denote bitrate savings.

\textbf{User Study {Settings}.}
To comprehensively assess perceptual quality, we conduct a user study following~\cite{Mentzer_22_ECCV_NVCGAN, Mentzer_2020_NeurIPS_HiFiC} at an ultra-low bitrate of approximately 0.002 bpp. We compare {Free-GVC} against the traditional codec VVC~\cite{Bross_2021_TCSVT_VVC}, the neural codec DCVC-RT~\cite{Jia_25_CVPR_DCVCRT}, and DCVC-RT{~\cite{Jia_25_CVPR_DCVCRT}} at double bitrate. {Additionally, the generative codec GLC-Video~\cite{Qi_25_TCSVT_GLCVideo} is included as a perceptual baseline at its minimum achievable bitrate, which is approximately three times that of our method}. An interactive web platform is developed where participants view the reference video alongside two randomly positioned comparison videos that play synchronously{, with one reconstructed by Free-GVC and the other by a competing method}. All reconstructed frames are exported as PNG and re-encoded into visually lossless H.265~\cite{Sullivan_2012_TCSVT_HEVC} videos using $\texttt{crf}=10$ at 20 fps for fair comparison. Eight sequences are randomly selected from HEVC B$\sim$E, UVG, and MCL-JCV datasets, resulting in 32 test groups. {A total of 23 participants aged 20--40 complete all test groups after receiving instructions and practice trials.} {For each comparison, participants indicate their preference based on overall perceptual quality, temporal continuity, and fidelity to the reference, with the comparison order and the positions of the two reconstructed videos randomly shuffled to mitigate potential ordering and positional bias. More details are provided in the supplementary material.}

\textbf{Implementation Details.}
We adopt CogVideoX-2B~\cite{Yang_25_ICLR_CogVideoX} as the generative prior, which natively supports 480p generation. {The model is operated in an unconditional manner by feeding an empty text prompt, i.e., no semantic guidance is provided during generation.} The VAE encoder and decoder are inherited from CogVideoX-2B{~\cite{Yang_25_ICLR_CogVideoX}}, with a temporal downsampling factor $s=4$, a spatial downsampling factor $d=8$, and a latent channel dimension of $C=16$. For high-resolution inputs, we employ a tiling strategy~\cite{Zhang_25_ICCV_StableCodec, Wang_24_IJCV_StableSR}, where each frame is partitioned into overlapping 480p tiles and processed independently by the VAE encoder and diffusion model. The resulting noisy latent tiles are aggregated using a Gaussian weighting map and jointly encoded via RCC to reduce spatial and temporal redundancy. At the decoder, noisy latents are partitioned, denoised, decoded by the VAE, and seamlessly stitched to reconstruct full-resolution frames. By default, videos are divided into GOPs of length $l=48$ with an overlap of $m=4$ frames. For inter-GOP alignment, we use a fixed weight $\gamma(i)=0.5$. Entropy coding is implemented with arithmetic coding. {
All BPP values reported in this paper are computed from the actual entropy-coded bitstreams, including the PFR seed indices and the diffusion timestep $t^*$ for each GOP. Bits from overlapping regions are also included, as the overlapped frames are explicitly encoded. The pseudo-random generator, diffusion model, and other codec configurations are fixed and shared between the encoder and decoder, and thus require no transmission.}

\begin{figure*}[t]
    \centering
    \includegraphics[width=1\linewidth]{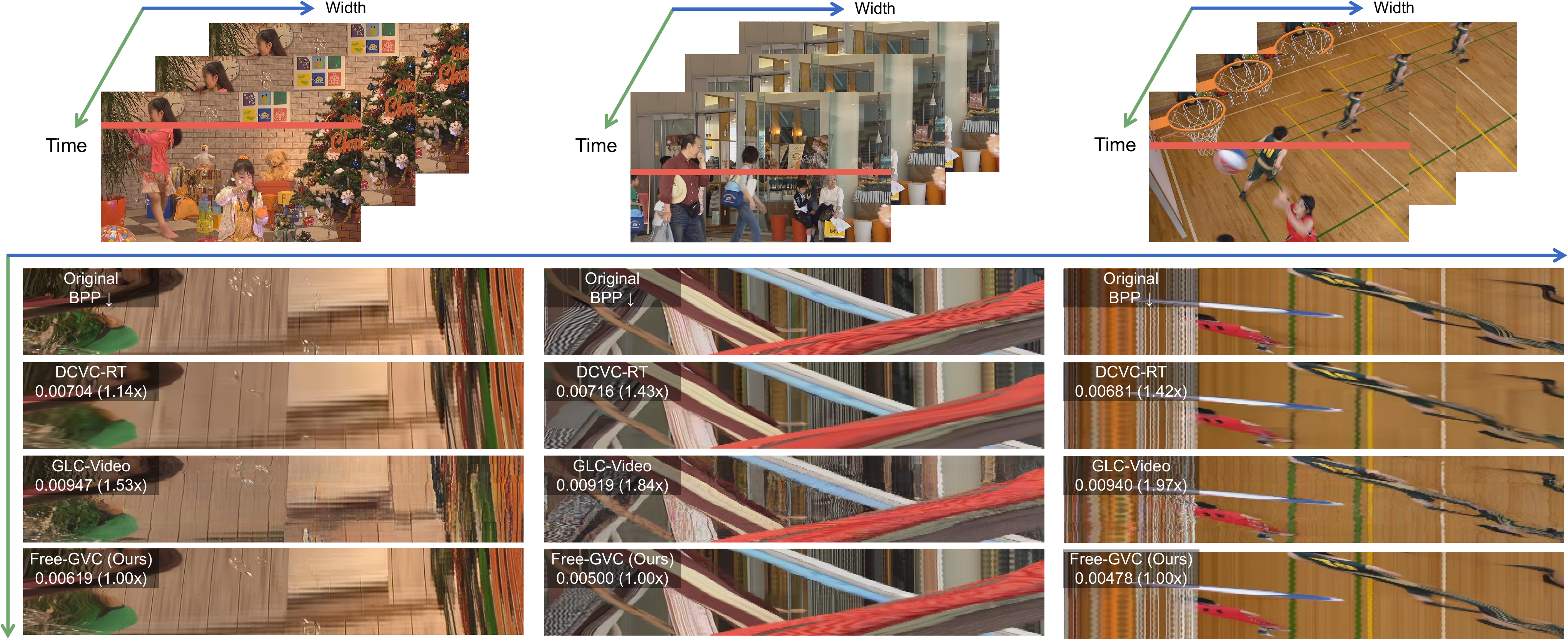}
    \caption{Visual comparison of temporal coherence by stacking the red line across {consecutive reconstructed frames from DCVC-RT~\cite{Jia_25_CVPR_DCVCRT}, GLC-Video~\cite{Qi_25_TCSVT_GLCVideo}, and the proposed Free-GVC}. Results that are closer to the original indicate better temporal coherence and quality. From left to right, the sequences are PartyScene, BQMall, and BasketballDrill from HEVC Class C. Zoom in for better view. }
    \label{continuity}
\end{figure*}

\subsection{Performance Comparison}
\textbf{Quantitative Results.}
Fig.~\ref{main} and Table~\ref{tab:main} summarize the rate--quality performance on HEVC Class B$\sim$E, UVG, and MCL-JCV datasets under different evaluation metrics. Overall, Free-GVC achieves the best perceptual performance across datasets, consistently outperforming all competing methods on the perception-oriented metrics DISTS and FID, demonstrating superior perceptual quality and distributional realism. For the distortion-oriented metric MS-SSIM, Free-GVC exhibits the expected perception--distortion trade-off associated with generative compression. Unlike conventional distortion-oriented codecs that primarily optimize pixel-level fidelity, Free-GVC leverages the generative prior of video diffusion models to enhance perceptual quality under extremely limited bitrates. Consequently, the reconstructed frames may deviate from the pixel-level references in certain regions, leading to lower MS-SSIM values in some cases. {A similar trend is observed for VMAF, which, as a full-reference metric, remains sensitive to deviations from the reference signal and therefore does not always favor Free-GVC over distortion-oriented codecs.} Importantly, this trade-off is more favorable than that of existing generative compression methods, such as GLC-Video{\cite{Qi_25_TCSVT_GLCVideo}}, DiffVC{\cite{Ma_25_tomm_DiffVC}}, and DiffC{\cite{Vonderfecht_25_ICLR_DiffC}}. Free-GVC not only achieves consistently superior performance on the perceptual metrics DISTS and FID, but also maintains stronger MS-SSIM {and VMAF} performance than these generative approaches. Overall, these results demonstrate that Free-GVC achieves a more favorable perception--distortion trade-off under ultra-low bitrate conditions.

\begin{figure}[t]
    \centering
    \includegraphics[width=0.95\linewidth]{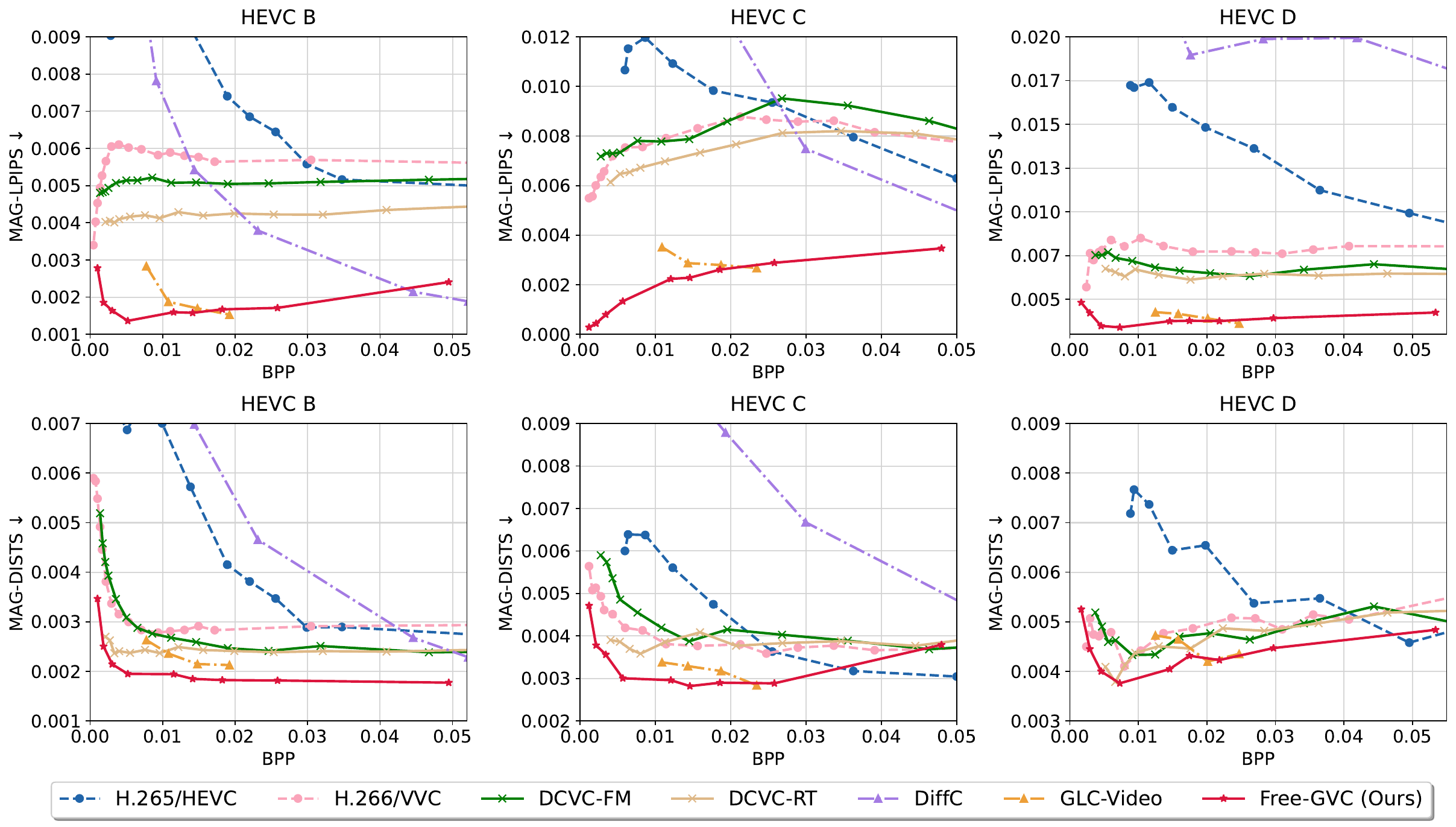}
    \caption{{Rate--MAG curves on the HEVC Class B--D datasets~\cite{Flynn_16_ITU_HEVC} for H.265/HEVC~\cite{Sullivan_2012_TCSVT_HEVC}, H.266/VVC~\cite{Bross_2021_TCSVT_VVC}, DCVC-FM~\cite{Li_24_CVPR_DCVCFM}, DCVC-RT~\cite{Jia_25_CVPR_DCVCRT}, DiffC~\cite{Vonderfecht_25_ICLR_DiffC}, GLC-Video~\cite{Qi_25_TCSVT_GLCVideo}, and the proposed Free-GVC. Lower MAG indicates smoother frame-wise quality variation.}}
    \label{fig:2_7_MAG}
\end{figure}

The advantages of Free-GVC are particularly pronounced on lower-resolution videos and at ultra-low bitrates. Under extreme compression, high-frequency details are inevitably discarded, and reconstruction quality is dominated by low-frequency structure modeling and semantic coherence, which is precisely the regime where our video diffusion prior excels. The results on HEVC Class C and D further corroborate this, and we provide additional compression results on low-resolution videos in the supplementary material to substantiate this observation. On 1080p sequences, since the video diffusion prior is trained on relatively low-resolution data, consistently modeling fine-grained perceptual details across tiles at large resolutions remains challenging, which leads to a slight gap in LPIPS compared to GLC-Video~\cite{Qi_25_TCSVT_GLCVideo}. Nevertheless, as prior works have noted~\cite{Qi_25_TCSVT_GLCVideo, Zhang_25_ICCV_StableCodec, Ding_22_TPAMI_DISTS}, LPIPS is less consistent with human perception than DISTS in compression scenarios. Given that Free-GVC achieves clear superiority on DISTS across all datasets and resolutions, we regard DISTS as the primary metric for perceptual comparison, and the overall results confirm the advantage of our method.

\begin{figure}[t]
    \centering
    \includegraphics[width=0.95\linewidth]{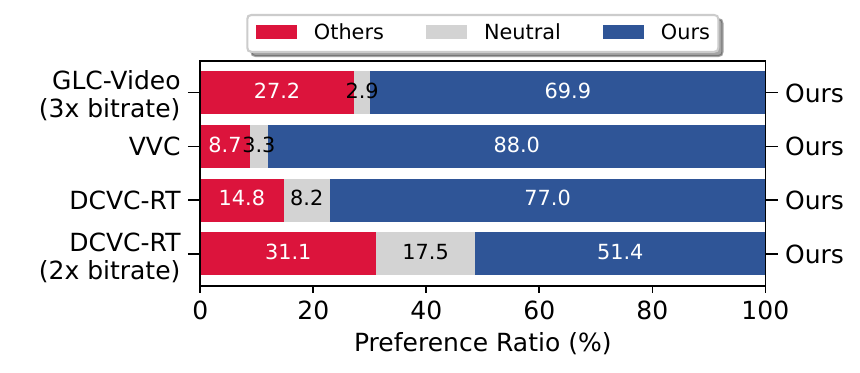}
    \caption{User study results on randomly selected videos. Our method consistently surpasses {VVC\cite{Bross_2021_TCSVT_VVC}, DCVC-RT\cite{Jia_25_CVPR_DCVCRT}, and GLC-Video\cite{Qi_25_TCSVT_GLCVideo}}, even when operating at a significantly lower bitrate.}
    \label{user}
\end{figure}

\begin{figure}[t]
    \centering
    \includegraphics[width=1\linewidth]{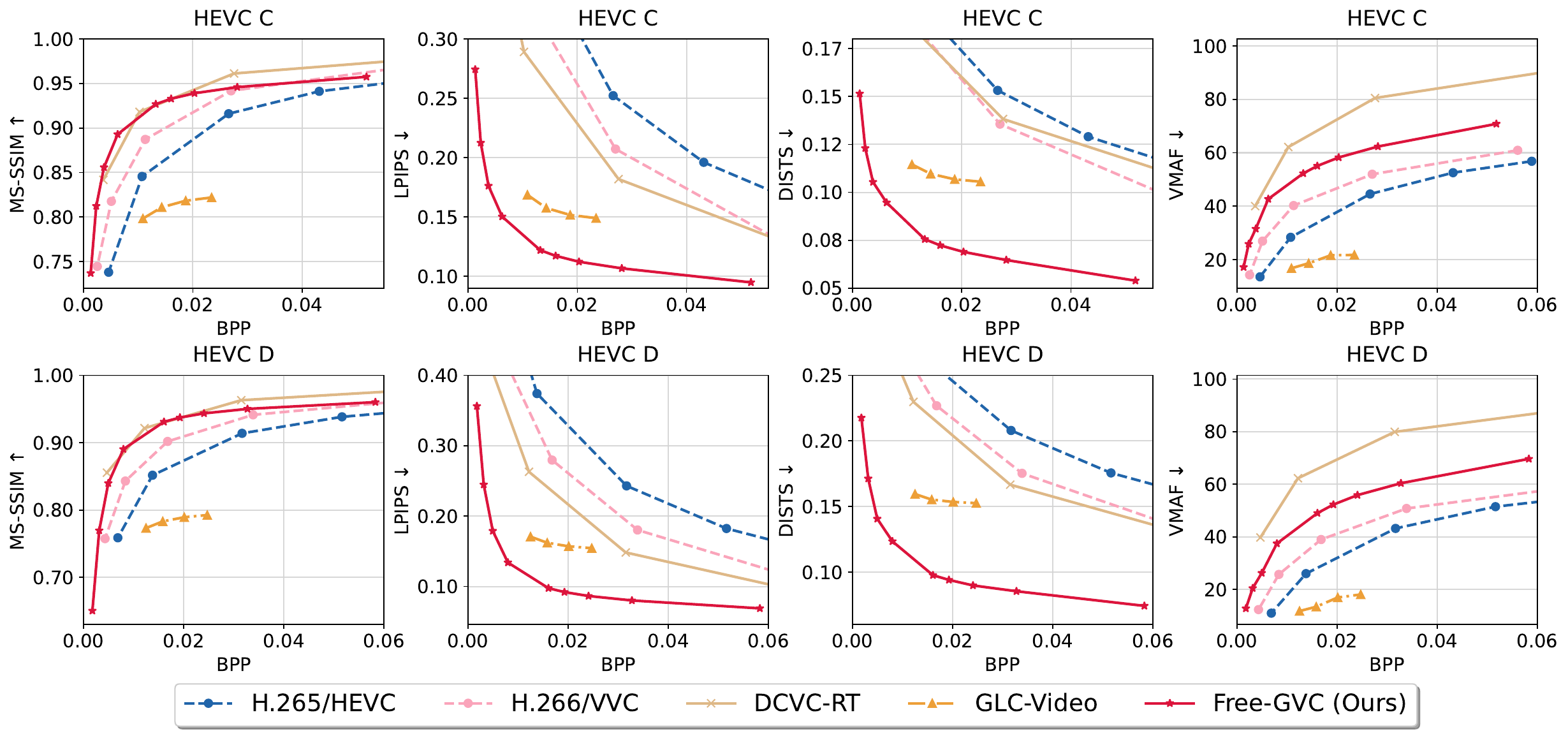}
    \caption{Rate--metric results {for H.265/HEVC~\cite{Sullivan_2012_TCSVT_HEVC}, H.266/VVC~\cite{Bross_2021_TCSVT_VVC}, DCVC-RT~\cite{Jia_25_CVPR_DCVCRT}, GLC-Video~\cite{Qi_25_TCSVT_GLCVideo}, and the proposed Free-GVC} on 400-frame sequences from HEVC Class C and D~\cite{Flynn_16_ITU_HEVC} for long-sequence evaluation.}
    \label{fig:1_3}
\end{figure}

\textbf{Qualitative Results.}  Fig.~\ref{visual} presents a visual comparison of reconstructions from different methods. At ultra-low bitrates, distortion-oriented codecs such as HEVC~\cite{Sullivan_2012_TCSVT_HEVC}, VVC~\cite{Bross_2021_TCSVT_VVC}, DCVC-FM~\cite{Li_24_CVPR_DCVCFM}, and DCVC-RT~\cite{Jia_25_CVPR_DCVCRT} exhibit noticeable blocking artifacts and motion ghosting. Perception-oriented generative methods like GLC-Video~\cite{Qi_25_TCSVT_GLCVideo} improve visual realism but may introduce spatial inconsistencies, producing visually plausible yet content-inaccurate details. In contrast, our approach leverages video diffusion prior to jointly model spatial and temporal redundancies, achieving ultra-low bitrate compression while preserving both perceptual fidelity and temporal coherence.

Fig.~\ref{continuity} further highlights temporal coherence. In regions with cross-frame object motion, DCVC-RT~\cite{Jia_25_CVPR_DCVCRT} produces heavy blur, while GLC-Video~\cite{Qi_25_TCSVT_GLCVideo} exhibits visible flickering on moving elements. Free-GVC, by contrast, maintains sharper details and smoother temporal transitions, demonstrating improved perceptual quality and temporal consistency. {To quantify temporal quality fluctuations, we adopt the Mean Absolute Gradient (MAG)~\cite{Fani_25_TOMM_MAG}, defined as
$\mathrm{MAG}(\delta)=\frac{1}{N_f-1}\sum_{i=1}^{N_f-1}|\delta_{i+1}-\delta_i|$,
where $\delta$ denotes a frame-wise quality metric. Lower MAG indicates smoother frame-wise quality variation. Figure~\ref{fig:2_7_MAG} presents the Rate--MAG curves on the HEVC Class B--D datasets. Free-GVC generally achieves lower MAG values in the ultra-low-bitrate range, indicating more stable temporal perceptual quality.
}

\textbf{User Study {Results}.}
As shown in Fig.~\ref{user}, Free-GVC is consistently preferred over both traditional and neural codecs at similar bitrate levels. Specifically, even when operating at roughly half the bitrate of DCVC-RT~\cite{Jia_25_CVPR_DCVCRT}, Free-GVC is preferred in 51.4\% of the comparisons. Furthermore, Free-GVC is preferred in 69.90\% of the comparisons against the generative codec GLC-Video~\cite{Qi_25_TCSVT_GLCVideo}, despite using only one-third of its bitrate. These results demonstrate that Free-GVC achieves superior perceptual quality and temporal coherence while maintaining substantially higher bitrate efficiency.

\begin{figure}[t]
    \centering
    \includegraphics[width=0.9\linewidth]{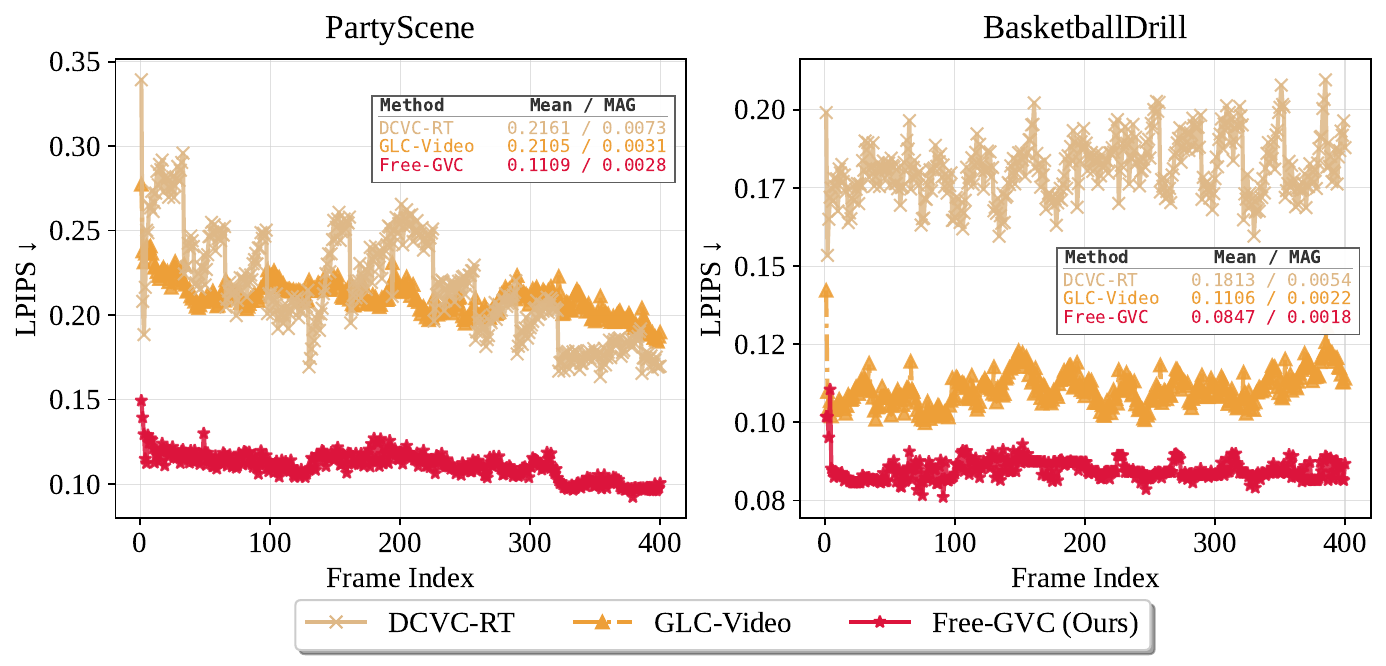}
    \caption{{Frame-wise quality metrics over 400-frame sequences from the HEVC Class C~\cite{Flynn_16_ITU_HEVC}. The BPP values for Free-GVC, GLC-Video~\cite{Qi_25_TCSVT_GLCVideo}, and DCVC-RT~\cite{Jia_25_CVPR_DCVCRT} are 0.0130, 0.0209, and 0.0154 for \textit{PartyScene}, and 0.0113, 0.0218, and 0.0186 for \textit{BasketballDrill}, respectively. }}
    \label{fig:2_7_framewise}
\end{figure}

\begin{table}[t]
\centering
\caption{Comparison with group-wise generative methods. BD-rate (\%) is computed using each method as the reference. Negative values indicate bitrate savings of Free-GVC.}
\label{tab:goplevel_comparison}
\begin{tabular}{c|cc}
\toprule
\textbf{Anchor Method} &
\textbf{LPIPS $\downarrow$} &
\textbf{DISTS $\downarrow$} \\
\midrule
T-GVC~\cite{Wang_25_arxiv_TGVC}
& -79.89\%
& -83.67\% \\
EVC-PDM~\cite{Li_24_WCSP_EVCDiffusion}
& -78.47\%
& -81.53\% \\
\bottomrule
\end{tabular}
\end{table}

{
\textbf{Long-Sequence Evaluation.}
To further assess the temporal stability of Free-GVC over extended video sequences, we conduct additional experiments on 400-frame sequences from HEVC Class C and Class D, which cover multiple GOPs and GOP boundaries. As shown in Fig.~\ref{fig:1_3}, Free-GVC maintains stable rate--quality performance over the complete sequences, demonstrating that the proposed framework remains effective beyond the original 96-frame evaluation setting. Furthermore, Fig.~\ref{fig:2_7_framewise} presents the frame-wise LPIPS curves over the 400-frame sequences, with the compared methods evaluated at comparable bitrate ranges. Free-GVC consistently achieves {lower mean LPIPS and MAG values} than GLC-Video{\cite{Qi_25_TCSVT_GLCVideo}} and DCVC-RT{\cite{Jia_25_CVPR_DCVCRT}}, indicating both better perceptual quality and more stable reconstruction over time. 

}

{
\textbf{Comparison with GOP-Level Generative Methods.}
To further evaluate the effectiveness of Free-GVC against existing GOP-level generative video compression approaches, we compare our method with two representative generative video compression methods, EVC-PDM~\cite{Li_24_WCSP_EVCDiffusion} and T-GVC~\cite{Wang_25_arxiv_TGVC}. Since these methods adopt different evaluation protocols, we follow their original settings for a fair comparison. Specifically, EVC-PDM{\cite{Li_24_WCSP_EVCDiffusion}} is evaluated on the first 30 frames of HEVC Class D sequences with the recommended input resolution of 128$\times$128, while T-GVC{\cite{Wang_25_arxiv_TGVC}} is evaluated on HEVC Class C sequences with 96 frames per sequence and a spatial resolution of 512$\times$320. The results are summarized in Table~\ref{tab:goplevel_comparison}. Free-GVC achieves more than 78\% bitrate savings across the evaluated perceptual metrics over both GOP-level generative baselines, demonstrating superior rate efficiency in the ultra-low bitrate regime.
}

\subsection{Ablation Study}

\begin{figure}[t]
    \centering
    \includegraphics[width=0.9\linewidth]{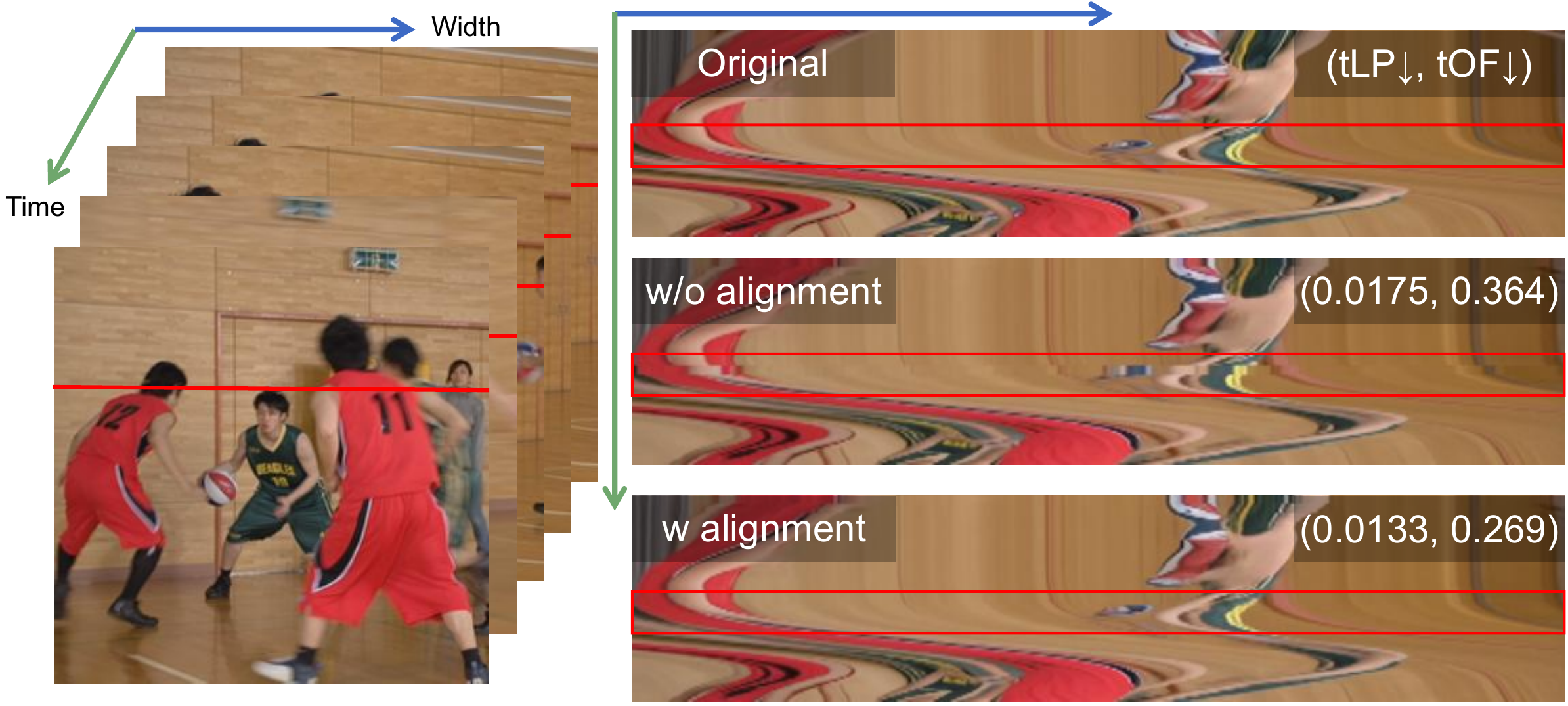}
    \caption{Comparison of temporal consistency with and without the Inter-GOP Alignment mechanism, visualized by stacking the red line over time. Zoom in for a clearer comparison.}
    \label{continue}
\end{figure}

\textbf{Inter-GOP Alignment.}
Fig.~\ref{continue} shows the temporal variation of pixel values at the same location before and after applying our alignment strategy. Without alignment, frames remain smooth within each GOP but exhibit artifacts at GOP boundaries. Inter-GOP alignment effectively reduces these inconsistencies, producing variations that closely follow the ground truth. We further examine the impact of the number of overlapping frames $m$ between adjacent GOPs in Table~\ref{tab:ablation_overlap_fusion}. A small overlap with $m=4$ slightly increases bitrate but improves temporal continuity and perceptual quality, often yielding comparable or better BD-Rate results, whereas enlarging the overlap to $m=8$ introduces redundant overhead and may amplify inconsistencies. Experiments on the fusion weight $\gamma(i)$ indicate that $\gamma(i)=0.5$ provides a favorable balance between temporal smoothness and reconstruction fidelity.

\begin{table}[t]
    \centering
    \renewcommand{\arraystretch}{1.15}
    \setlength{\tabcolsep}{5pt}
    \caption{{Ablation study on the number of overlap frames $m$ and
    the fusion weight $\gamma(i)$.
    Ours uses $m=4$ and $\gamma(i)=0.5$.
    Each ablation varies one parameter while keeping the other fixed.
    LPIPS and DISTS are reported as BD-Rate (\%) relative to Ours.
    Bold indicates the best result within each ablation.}}
    \label{tab:ablation_overlap_fusion}
    \footnotesize
    \begin{tabular}{c cc c ccc}
        \toprule
        \multirow{2}{*}{\textbf{Metric}}
        & \multicolumn{2}{c}{\textbf{Overlap Frames $m$}}
        & \multirow{2}{*}{\textbf{Ours}}
        & \multicolumn{3}{c}{\textbf{Fusion Weight $\gamma(i)$}} \\
        \cmidrule(lr){2-3}
        \cmidrule(lr){5-7}
        & 0 & 8 & & 0.4 & 0.6 & 0.7 \\
        \midrule
        LPIPS $\downarrow$
            & 8.68 & 33.91
            & \textbf{0.00}
            & 7.43 & 3.25 & 8.14 \\
        DISTS $\downarrow$
            & \textbf{-0.93} & 44.39
            & 0.00
            & \textbf{-6.04} & 4.32 & 1.83 \\
        \midrule
        tLP $\downarrow$
            & 0.0254 & 0.0214
            & \textbf{0.0199}
            & 0.0201 & 0.0201 & 0.0201 \\
        tOF $\downarrow$
            & 0.3960 & 0.3654
            & \textbf{0.3061}
            & 0.3076 & 0.3108 & 0.3070 \\
        \bottomrule
    \end{tabular}
\end{table}

\begin{figure}[t]
    \centering
    \includegraphics[width=0.9\linewidth]{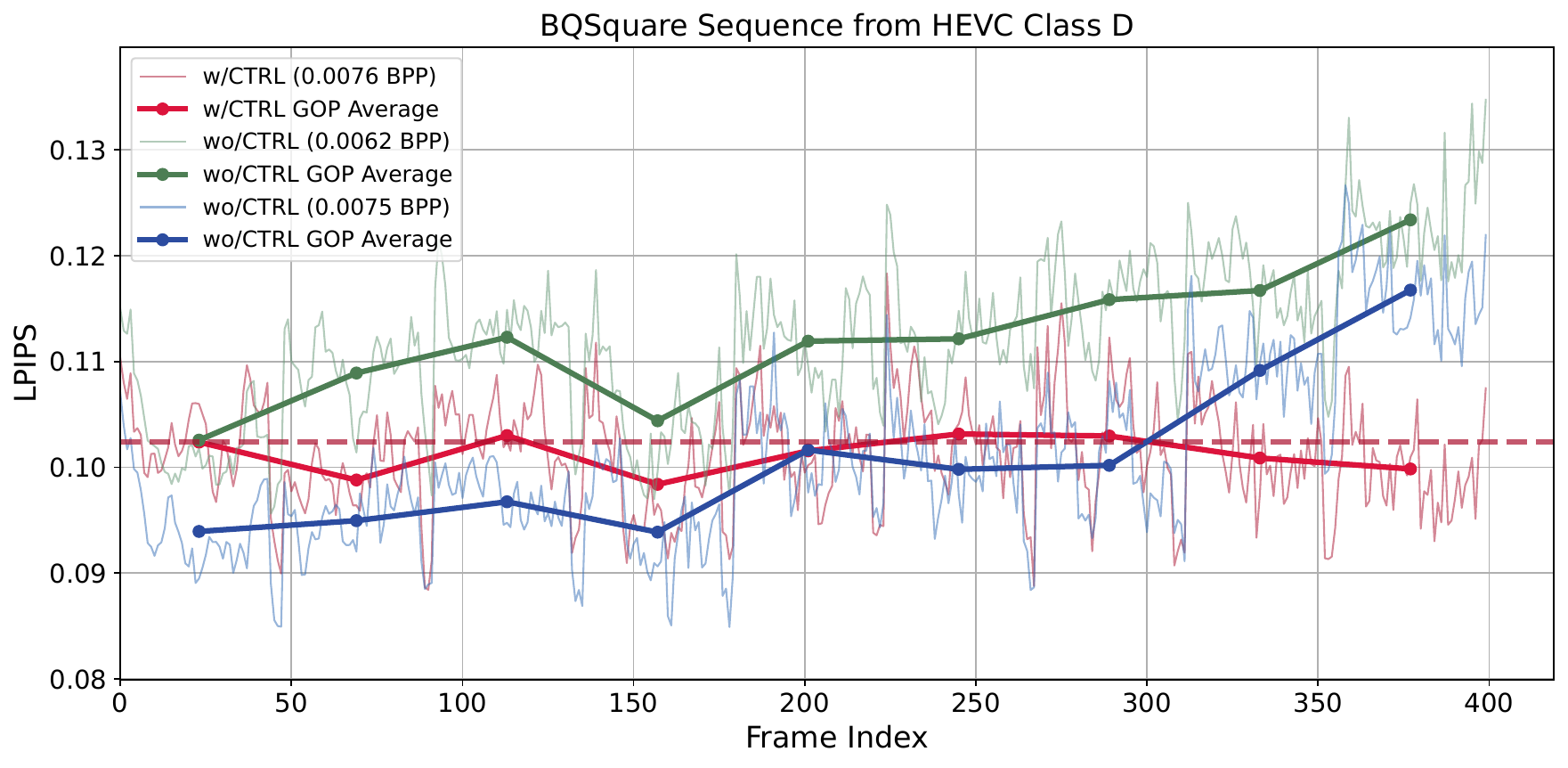}
    \caption{LPIPS variation across frame indexes, comparing results with and without the proposed quality control algorithm.}
    \label{control}
\end{figure}

\textbf{Quality Control.}
Fig.~\ref{control} shows the effect of our quality control across GOPs. We evaluate on the first 400 frames of each video. Without this mechanism, content variations cause large fluctuations in reconstruction quality. In contrast, our method uses the quality metric of the first GOP as a target for subsequent GOPs, dynamically adjusting the encoding process to maintain stable perceptual quality. With prediction reuse, only two to three decoding iterations are typically required for accurate quality estimation, compared with five to seven without reuse. {We further investigate the impact of the quality-control threshold $\epsilon$. As shown in Table~\ref{tab:threshold}, a smaller $\epsilon$ improves quality matching but requires more refinement steps. We select $\epsilon=0.005$ as a good trade-off, achieving a quality deviation of 0.0045 with only 3.0 refinement steps on average.}

\begin{figure}[!t]
    \centering
    \includegraphics[width=0.8\linewidth]{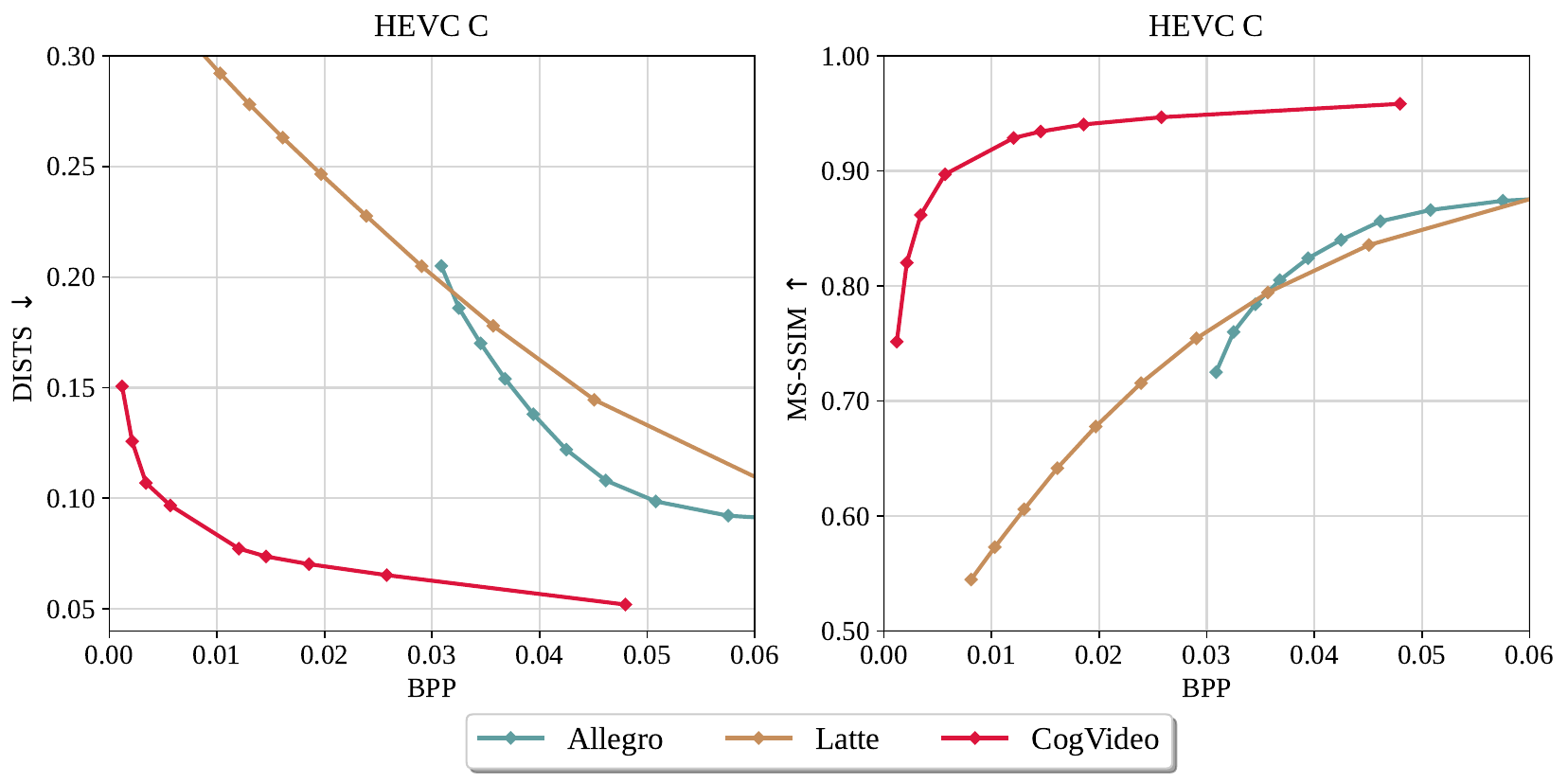}
    \caption{{Rate--quality results for Free-GVC using CogVideoX-2B~\cite{Yang_25_ICLR_CogVideoX}, Latte~\cite{Ma_25_TMLR_Latte}, and Allegro~\cite{Zhou_24_arxiv_allegro} on HEVC Class C~\cite{Flynn_16_ITU_HEVC}.}}
    \label{fig:backbone_main}
\end{figure}

{
\textbf{Cross-Backbone and GOP-Size Analysis.}
Free-GVC is not tied to a specific pretrained video generative model. To evaluate its generality and justify the default backbone, we replace CogVideoX-2B~\cite{Yang_25_ICLR_CogVideoX} with Latte~\cite{Ma_25_TMLR_Latte} and Allegro~\cite{Zhou_24_arxiv_allegro}, while keeping the remaining compression pipeline unchanged. Fig.~\ref{fig:backbone_main} shows that all three video diffusion models can serve as generative backbones, with CogVideoX-2B achieving the best overall performance and thus being adopted as the default backbone in the experiments. We further investigate the GOP-length sensitivity of each backbone in Table~\ref{tab:backbone_gop}, using its best GOP configuration as the BD-Rate anchor. The preferred GOP sizes are 48, 24, and 56 frames for CogVideoX-2B\cite{Yang_25_ICLR_CogVideoX}, Latte\cite{Ma_25_TMLR_Latte}, and Allegro\cite{Zhou_24_arxiv_allegro}, respectively. Notably, the preferred GOP sizes of Latte and Allegro differ from their native generation lengths of 16 and 88 frames, respectively, indicating that the optimal GOP size is determined by the interaction between the backbone and compression setting rather than simply by its native temporal length.
}

\begin{table}[t]
\centering
\caption{{Effect of different thresholds $\epsilon$ in the quality control module. \emph{Quality deviation} denotes the average absolute error between the target and achieved perceptual quality, while \emph{Refinement steps} denotes the average number of iterative refinement steps. Lower values are better for both metrics.}}
\label{tab:threshold}
\begin{tabular}{c|c|c}
\hline
\textbf{Threshold $\epsilon$} & \textbf{Quality deviation $\downarrow$} & \textbf{Refinement steps} $\downarrow$ \\
\hline
0.001 & 0.0021 & 7.0 \\
0.005 & 0.0045 & 3.0 \\
0.010 & 0.0078 & 2.3 \\
0.020 & 0.0136 & 2.0 \\
\hline
\end{tabular}
\end{table}

\begin{table}[!t]
    \centering
    \caption{{LPIPS/DISTS BD-Rate (\%) on HEVC Class C~\cite{Flynn_16_ITU_HEVC} 
    with different GOP sizes and pretrained video diffusion backbones. 
    The best-performing GOP is used as the anchor for each backbone.}}
    \label{tab:backbone_gop}
    \renewcommand{\arraystretch}{1.15}
    \setlength{\tabcolsep}{4pt}

    \begin{tabular}{c|ccc}
    \toprule
    \textbf{GOP Size} &
    \textbf{CogVideoX-2B\cite{{Yang_25_ICLR_CogVideoX}}} &
    \textbf{Latte\cite{Ma_25_TMLR_Latte}} &
    \textbf{Allegro\cite{Zhou_24_arxiv_allegro}} \\
    \midrule
    16 & 173.54/263.57 & 24.83/28.56 & -- \\
    24 & 78.45/84.24   & \textbf{0/0} & -- \\
    32 & 36.39/39.75   & 7.46/8.21    & 32.67/36.84 \\
    48 & \textbf{0/0}  & 8.92/9.37    & 18.62/21.47 \\
    56 & 6.75/6.87     & --           & \textbf{0/0} \\
    64 & --             & --           & 4.73/5.38 \\
    \bottomrule
    \end{tabular}
\end{table}

\subsection{Complexity Analysis}

\begin{table}[t]
    \centering
    \renewcommand{\arraystretch}{1.2}
    \caption{Detailed runtime of various modules on HEVC C dataset at 480p using NVIDIA RTX 4090.}
    \label{tab:time}

    \resizebox{\columnwidth}{!}{
    \begin{tabular}{c|cc|ccc|c}
    \toprule
    \textbf{Timestep} & \textbf{MS-SSIM $\uparrow$} & \textbf{DISTS $\downarrow$} & 
    \textbf{VAE Enc.} & \textbf{Diffusion} & \textbf{RCC} &
    \textbf{Enc. FPS}  \\
    \midrule
    50 & 0.9594 & 0.0711 & 0.0555 & 0.522 & 3.349 & 0.254 \\
    100 & 0.9503 & 0.0764 & 0.0555 & 0.436 & 2.598 & 0.323   \\
    200 & 0.9280 & 0.0966 & 0.0555 & 0.293 & 1.564 & 0.522   \\
    300 & 0.9047 & 0.0986 & 0.0555 & 0.221 & 1.118 & 0.716   \\
    400 & 0.8738 & 0.112 & 0.0555 & 0.172 & 0.844 & 0.932\\
    500 & 0.8230 & 0.133 & 0.0555 & 0.123 & 0.543 & 1.38\\
    \midrule
    \multicolumn{7}{c}{\textit{(a) Encoding time (seconds) and FPS}} \\
    \bottomrule
    \end{tabular}
    }
    
    \vspace{0.3cm}

    \resizebox{\columnwidth}{!}{
    \begin{tabular}{c|c|cccc|c}
    \toprule
    \textbf{Timestep} & \textbf{BPP $\downarrow$} &
    \textbf{Diffusion} & \textbf{RCC} & \textbf{Denoise} & \textbf{VAE Dec.} & \textbf{Dec. FPS} \\
    \midrule
    50 & 0.0152 & 0.507 & 2.48 & 0.0228 & 0.111  & 0.319 \\
    100 & 0.0099 & 0.422 & 1.968 & 0.0318 & 0.113  & 0.394 \\
    200 & 0.0047 & 0.278 & 1.24 & 0.0591 & 0.109  & 0.589 \\
    300 & 0.0028 & 0.206 & 0.912 & 0.0773 & 0.111  & 0.764 \\
    400 & 0.0018 & 0.157 & 0.678 & 0.100 & 0.112  & 0.954 \\
    500 & 0.0011 & 0.107 & 0.459 & 0.123 & 0.109  & 1.250 \\
    \midrule
    \multicolumn{7}{c}{\textit{(b) Decoding time (seconds) and FPS}} \\
    \bottomrule
    \end{tabular}
    }
\end{table}

\begin{table}[!t]
\centering
\caption{{Computational complexity comparison at $832\times480$ resolution. Available implementations were measured on an NVIDIA RTX 5090. The DiffVC result marked with $\ast$ is quoted from its publication and was measured on an NVIDIA A800. BD-Rate is computed from DISTS with x265 as the anchor.}}
\label{tab:complexity}

\resizebox{\columnwidth}{!}{%
\begin{tabular}{@{}c|c|c|c|c@{}}
\toprule
\textbf{Method}
& \textbf{Enc./Dec. FPS}
& \textbf{kMACs/pixel}
& \textbf{Params (M)}
& \textbf{BD-Rate(\%)} \\
\midrule

x265-3.3~\cite{Sullivan_2012_TCSVT_HEVC}
& 26.07 / 51.91
& --
& --
& 0.00 \\

DCVC-FM~\cite{Li_24_CVPR_DCVCFM}
& 19.06 / 22.32
& 1128.62
& 44.93
& -8.99 \\

DCVC-RT~\cite{Jia_25_CVPR_DCVCRT}
& 113.99 / 123.13
& 205.26
& 66.34
& -11.42 \\

GLC-Video~\cite{Qi_25_TCSVT_GLCVideo}
& 38.92 / 23.19
& 3170.24
& 289.33
& -75.72 \\

DiffVC$^\ast$~\cite{Ma_25_tomm_DiffVC}
& 0.29 / 0.29
& 110083.01
& 1686.07
& N/A \\

Free-GVC
& 1.40 / 1.28
& 74038.15
& 1909.37
& -94.81 \\

\bottomrule
\end{tabular}%
}
\end{table}

Table~\ref{tab:time} summarizes the encoding and decoding runtime of Free-GVC. As a progressive coding framework, higher target bitrates require more diffusion and RCC steps, leading to a near-linear increase in runtime, with RCC scaling more rapidly than the diffusion computations. While real-time processing remains challenging, further acceleration may be achieved through RCC optimization~\cite{Ohayon_25_ICML_DDCM, Vaisman_25_arxiv_turboddcm} and diffusion speed-up techniques~\cite{Ma_24_CVPR_DeepCache, Fang_23_NIPS_prune, Fang_25_CVPR_Tinyfusion}. {
Table~\ref{tab:complexity} compares the computational complexity and compression performance at $832\times480$ resolution on HEVC Class C. Although Free-GVC requires more computation than conventional and neural codecs, it provides a substantial perceptual compression advantage, achieving a DISTS BD-Rate of $-94.81\%$ relative to x265~\cite{Sullivan_2012_TCSVT_HEVC}. Moreover, its kMACs/pixel is substantially lower than that of DiffVC~\cite{Ma_25_tomm_DiffVC}. These results demonstrate that Free-GVC offers a favorable trade-off between computational cost and perceptual compression performance while improving efficiency over prior diffusion-based video codecs.
}

\section{Conclusion}
\label{sec:conclusion}
This paper presents a {retraining-free} generative video compression framework, termed Free-GVC, which leverages pre-trained video diffusion models to jointly capture both spatial and temporal dependencies. By reformulating compression as progressive latent coding within the diffusion space, our method achieves perceptually faithful and visually coherent reconstruction at ultra-low bitrates. The Inter-GOP Alignment module effectively mitigates boundary artifacts and enhances temporal consistency, while the Adaptive Quality Control mechanism dynamically stabilizes perceptual quality across diverse content and GOPs.

Despite its strong performance, achieving real-time compression for high-resolution videos remains challenging for this framework. Nevertheless, this limitation can be alleviated through future optimization, highlighting a promising direction for generative video compression.

\bibliographystyle{IEEEtran}
\bibliography{references}

@String(PAMI   = {IEEE Trans. Pattern Anal. Mach. Intell.})

@String(IJCV   = {Int. J. Comput. Vis.})

@String(TIP    = {IEEE Trans. Image Process.})

@String(TMM    = {IEEE Trans. Multimedia})

@String(TCSVT  = {IEEE Trans. Circuits Syst. Video Technol.})

@string(TBC    = {IEEE Trans. Broadcast.})

@String(CVPR   = {Proc. IEEE/CVF Conf. Comput. Vis. Pattern Recognit.})

@String(ICCV   = {Proc. IEEE/CVF Int. Conf. Comput. Vis.})

@String(ECCV   = {Proc. Eur. Conf. Comput. Vis.})

@String(NeurIPS= {Adv. Neural Inf. Process. Syst.})

@String(ICLR   = {Int. Conf. Learn. Represent.})

@String(ICML   = {Int. Conf. Mach. Learn.})

@String(AAAI   = {Proc. AAAI Conf. Artif. Intell.})

@String(IJCAI  = {Proc. Int. Joint Conf. Artif. Intell.})

@String(ACMMM  = {Proc. ACM Int. Conf. Multimedia})

@String(ICASSP = {Proc. IEEE Int. Conf. Acoust. Speech Signal Process.})

@String(ICIP   = {Proc. IEEE Int. Conf. Image Process.})

@String(ASILOMAR = {Proc. Asilomar Conf. Signals Syst. Comput.})

@String(MMSYS = {Proc. ACM Multimedia Syst. Conf.})

@String(TOMM = {ACM Trans. Multimedia Comput. Commun. Appl.})

@String(WCSP = {Proc. Int. Conf. Wireless Commun. Signal Process.})

@String(DCC = {Proc. Data Compression Conf.})

@String(CCC = {Proc. IEEE Conf. Comput. Complexity})

@String(VCIP = {Proc. IEEE Vis. Commun. Image Process.})

@String(TIT = {IEEE Trans. Inf. Theory})

@String(VI = {Visual Intell.})

@String(TMLR = {Transactions on Machine Learning Research})

@article{Bross_2021_TCSVT_VVC,
  author   = {Benjamin Bross and Ye-Kui Wang and Yan Ye and Shan Liu and Jianle Chen and Gary J. Sullivan and Jens-Rainer Ohm},
  title    = {Overview of the Versatile Video Coding (VVC) Standard and its Applications},
  journal  = TCSVT,
  year     = {2021},
  volume   = {31},
  number   = {10},
  pages    = {3736-3764}
}

@inproceedings{Careil_2023_ICLR_Perco,
  title     = {Towards image compression with perfect realism at ultra-low bitrates},
  author    = {Careil, Marl{\`e}ne and Muckley, Matthew J and Verbeek, Jakob and Lathuili{\`e}re, St{\'e}phane},
  booktitle = ICLR,
  year      = {2023}
}

@article{Goodfellow_2014_NeurIPS_GAN,
  title   = {Generative adversarial nets},
  author  = {Goodfellow, Ian and Pouget-Abadie, Jean and Mirza, Mehdi and Xu, Bing and Warde-Farley, David and Ozair, Sherjil and Courville, Aaron and Bengio, Yoshua},
  journal = NeurIPS,
  volume  = {27},
  year    = {2014}
}

@inproceedings{Heusel_2017_NeurIPS_FID,
  author    = {Martin Heusel and
               Hubert Ramsauer and
               Thomas Unterthiner and
               Bernhard Nessler and
               Sepp Hochreiter},
  title={GANs trained by a two time-scale update rule converge to a local Nash equilibrium},
  booktitle = NeurIPS,
  pages     = {6626--6637},
  year      = {2017}
}

@inproceedings{Ho_2020_NeurIPS_DDPM,
  author    = {Jonathan Ho and
               Ajay Jain and
               Pieter Abbeel},
  title     = {Denoising diffusion probabilistic models},
  booktitle = NeurIPS,
  year      = {2020}
}

@inproceedings{Li_2023_CVPR_DCVCDC,
  author    = {Jiahao Li and
               Bin Li and
               Yan Lu},
  title     = {Neural Video Compression with Diverse Contexts},
  booktitle = CVPR,
  pages     = {22616--22626},
  publisher = {},
  year      = {2023}
}

@inproceedings{Lu_2019_CVPR_DVC,
  author    = {Guo Lu and
               Wanli Ouyang and
               Dong Xu and
               Xiaoyun Zhang and
               Chunlei Cai and
               Zhiyong Gao},
  title     = {{DVC:} An End-To-End Deep Video Compression Framework},
  booktitle = CVPR,
  pages     = {11006--11015},
  publisher = {},
  year      = {2019}
}

@inproceedings{Mentzer_2020_NeurIPS_HiFiC,
  author    = {Fabian Mentzer and
               George Toderici and
               Michael Tschannen and
               Eirikur Agustsson},
  title={High-fidelity generative image compression},
  booktitle = NeurIPS,
  year      = {2020}
}

@inproceedings{Podell_2024_ICLR_SDXL,
  author    = {Dustin Podell and
               Zion English and
               Kyle Lacey and
               Andreas Blattmann and
               Tim Dockhorn and
               Jonas M{\"{u}}ller and
               Joe Penna and
               Robin Rombach},
  title     = {{SDXL:} Improving Latent Diffusion Models for High-Resolution Image Synthesis},
  booktitle = ICLR,
  publisher = {},
  year      = {2024}
}

@article{Sullivan_2012_TCSVT_HEVC,
  author  = {Gary J. Sullivan and
             Jens{-}Rainer Ohm and
             Woojin Han and
             Thomas Wiegand},
  title   = {Overview of the High Efficiency Video Coding {(HEVC)} Standard},
  journal = TCSVT,
  volume  = {22},
  number  = {12},
  pages   = {1649--1668},
  year    = {2012}
}

@article{Wang_24_IJCV_StableSR,
  author       = {Jianyi Wang and
                  Zongsheng Yue and
                  Shangchen Zhou and
                  Kelvin C. K. Chan and
                  Chen Change Loy},
  title        = {Exploiting Diffusion Prior for Real-World Image Super-Resolution},
  journal      = IJCV,
  volume       = {132},
  number       = {12},
  pages        = {5929--5949},
  year         = {2024}
}

@inproceedings{Zhang_2018_CVPR_LPIPS,
  author    = {Richard Zhang and
               Phillip Isola and
               Alexei A. Efros and
               Eli Shechtman and
               Oliver Wang},
  title     = {The Unreasonable Effectiveness of Deep Features as a Perceptual Metric},
  booktitle = CVPR,
  pages     = {586--595},
  publisher = {},
  year      = {2018}
}

@inproceedings{Li_21_NeurIPS_DCVC,
  author       = {Jiahao Li and
                  Bin Li and
                  Yan Lu},
  title        = {Deep Contextual Video Compression},
  booktitle    = NeurIPS,
  pages        = {18114--18125},
  year         = {2021}
}

@inproceedings{Li_24_CVPR_DCVCFM,
  author       = {Jiahao Li and
                  Bin Li and
                  Yan Lu},
  title        = {Neural Video Compression with Feature Modulation},
  booktitle    = CVPR,
  pages        = {26099--26108},
  publisher    = {{IEEE}},
  year         = {2024}
}

@inproceedings{Yang_22_IJCAI_PLVC,
  author       = {Ren Yang and
                  Radu Timofte and
                  Luc Van Gool},
  title        = {Perceptual Learned Video Compression with Recurrent Conditional {GAN}},
  booktitle    = IJCAI,
  pages        = {1537--1544},
  year         = {2022}
}

@inproceedings{Li_23_MM_HVFVC,
  author       = {Meng Li and
                  Yibo Shi and
                  Jing Wang and
                  Yunqi Huang},
  title        = {High Visual-Fidelity Learned Video Compression},
  booktitle    = ACMMM,
  pages        = {8057--8066},
  publisher    = {},
  year         = {2023}
}

@ARTICLE{Li_25_TCSVT_DiffEIC,
  author={Zhiyuan Li and Yanhui Zhou and Hao Wei and Chenyang Ge and Jingwen Jiang},
  title={Toward Extreme Image Compression With Latent Feature Guidance and Diffusion Prior}, 
  journal=TCSVT, 
  year={2025},
  volume={35},
  number={1},
  pages={888-899}
}

@inproceedings{Yang_25_ICLR_CogVideoX,
  author       = {Zhuoyi Yang and
                  Jiayan Teng and
                  Wendi Zheng and
                  et al.},
  title={CogVideoX: Text-to-video diffusion models with an expert transformer},
  booktitle = ICLR,
  publisher = {},
  year      = {2025}
}

@article{Flynn_16_ITU_HEVC,
  title={{Common Test Conditions and Software Reference Configurations for HEVC Range Extensions, document JCTVC-N1006}},
  author={Flynn, D and Sharman, K and Rosewarne, C},
  journal={Joint Collaborative Team Video Coding ITU-T SG},
  volume={16},
  year={2011} 
}

@inproceedings{Mercat_20_MMsys_UVG,
  title={{UVG dataset: 50/120fps 4K sequences for video codec analysis and development}},
  author={Mercat, Alexandre and Viitanen, Marko and Vanne, Jarno},
  booktitle=MMSYS,
  pages={297--302},
  year={2020}
}

@inproceedings{Wang_16_ICIP_MCL-JCV,
  title={{MCL-JCV: a JND-based H. 264/AVC video quality assessment dataset}},
  author={Wang, Haiqiang and Gan, Weihao and Hu, Sudeng and Lin, Joe Yuchieh and Jin, Lina and Song, Longguang and Wang, Ping and Katsavounidis, Ioannis and Aaron, Anne and Kuo, C-C Jay},
  booktitle=ICIP,
  pages={1509--1513},
  year={2016},
  organization={IEEE}
}

@ARTICLE{Ding_22_TPAMI_DISTS,
  author={Ding, Keyan and Ma, Kede and Wang, Shiqi and Simoncelli, Eero P.},
  journal=PAMI, 
  title={Image Quality Assessment: Unifying Structure and Texture Similarity}, 
  year={2022},
  volume={44},
  number={5},
  pages={2567-2581}
}

@inproceedings{Jia_25_CoD,
  title={Cod: A diffusion foundation model for image compression},
  author={Jia, Zhaoyang and Zheng, Zihan and Xue, Naifu and Li, Jiahao and Li, Bin and Guo, Zongyu and Zhang, Xiaoyi and Li, Houqiang and Lu, Yan},
  booktitle=CVPR,
  pages={38420--38429},
  year={2026}
}

@inproceedings{Oh_22_ECCV_DeMFI,
  title={DeMFI: Deep Joint Deblurring and Multi-Frame Interpolation with Flow-Guided Attentive Correlation and Recursive Boosting},
  author={Oh, Jihyong and Kim, Munchurl},
  booktitle=ECCV,
  year={2022}
}

@InProceedings{Youk_2024_CVPR_FMANet,
    author    = {Youk, Geunhyuk and Oh, Jihyong and Kim, Munchurl},
    title     = {FMA-Net: Flow-Guided Dynamic Filtering and Iterative Feature Refinement with Multi-Attention for Joint Video Super-Resolution and Deblurring},
    booktitle = CVPR,
    month     = {June},
    year      = {2024},
    pages     = {44-55}
}

@inproceedings{Kroeger_16_ECCV_DIS, 
    Author    = {Till Kroeger and Radu Timofte and Dengxin Dai and Luc Van Gool}, 
    Title     = {Fast Optical Flow using Dense Inverse Search}, 
    Booktitle = ECCV, 
    Year      = {2016}
}

@inproceedings{Jia_24_CVPR_GLC,
  author       = {Zhaoyang Jia and
                  Jiahao Li and
                  Bin Li and
                  Houqiang Li and
                  Yan Lu},
  title={Generative latent coding for ultra-low bitrate image compression},
  booktitle    = CVPR,
  pages        = {26088--26098},
  publisher    = {},
  year         = {2024}
}

@article{Tim_24_Openai_Sora,
  title={Video generation models as world simulators},
  author={Tim Brooks and Bill Peebles and Connor Holmes and Will DePue and Yufei Guo and Li Jing and David Schnurr and Joe Taylor and Troy Luhman and Eric Luhman and Clarence Ng and Ricky Wang and Aditya Ramesh},
  year={2024},
  journal={},
  url={https://openai.com/research/video-generation-models-as-world-simulators},
}

@inproceedings{Li_16_Netflix_VMAF,
    author = {Zhi Li and Anne Aaron and Ioannis Katsavounidis and Anush Moorthy and Megha Manohara},
    title = {Toward A Practical Perceptual Video Quality Metric},
    booktitle = {Netflix Technology Blog},
    year = {2016}
}

@inproceedings{Esser_21_CVPR_VQGAN,
  author       = {Patrick Esser and
                  Robin Rombach and
                  Bj{\"{o}}rn Ommer},
  title        = {Taming transformers for high-resolution image synthesis},
  booktitle    = CVPR,
  pages        = {12873--12883},
  publisher    = {},
  year         = {2021}
}

@inproceedings{Jia_25_CVPR_DCVCRT,
  author       = {Zhaoyang Jia and
                  Bin Li and
                  Jiahao Li and
                  Wenxuan Xie and
                  Linfeng Qi and
                  Houqiang Li and
                  Yan Lu},
  title        = {Towards Practical Real-Time Neural Video Compression},
  booktitle    = CVPR,
  pages        = {12543--12552},
  year         = {2025}
}

@article{Pan_24_TBC_JND-LIC,
  title={JND-LIC: Learned image compression via just noticeable difference for human visual perception},
  author={Pan, Zhaoqing and Zhang, Guoyu and Peng, Bo and Lei, Jianjun and Xie, Haoran and Wang, Fu Lee and Ling, Nam},
  journal=TBC,
  volume={71},
  number={1},
  pages={217--228},
  year={2024},
  publisher={IEEE}
}

@article{Guo_TBC_2025_LVC,
  title={Exploring invertible encoding for deep video compression},
  author={Guo, Haifeng and Kwong, Sam and Zhou, Mingliang},
  journal=TBC,
  year={2025},
  publisher={IEEE}
}

@article{Yuan_TBC_2025_LVC,
  title={Multi-Domain Spatial-Temporal Redundancy Mining for Efficient Learned Video Compression},
  author={Yuan, Feng and Pan, Zhaoqing and Lei, Jianjun and Peng, Bo and Wang, Fu Lee and Kwong, Sam},
  journal=TBC,
  volume={71},
  number={3},
  pages={808--820},
  year={2025},
  publisher={IEEE}
}

@article{Zhaoyan_TBC_2024_Video,
  title={An efficient and flexible complexity control method for versatile video coding},
  author={Zhao, Yan and Zhu, Chen and Xu, Jun and Lu, Guo and Song, Li and Ma, Siwei},
  journal=TBC,
  volume={71},
  number={1},
  pages={96--110},
  year={2024},
  publisher={IEEE}
}

@article{Wangshen_TBC_2024_Video,
  title={Long-term and short-term information propagation and fusion for learned video compression},
  author={Wang, Shen and Feng, Donghui and Lu, Guo and Cheng, Zhengxue and Song, Li and Zhang, Wenjun},
  journal=TBC,
  volume={70},
  number={4},
  pages={1254--1265},
  year={2024},
  publisher={IEEE}
}

@article{zhao_TBC_2023_bg,
  title={End-to-end compression for surveillance video with unsupervised foreground-background separation},
  author={Zhao, Yu and Luo, Dengyan and Wang, Fuchun and Gao, Han and Ye, Mao and Zhu, Ce},
  journal=TBC,
  volume={69},
  number={4},
  pages={966--978},
  year={2023},
  publisher={IEEE}
}

@inproceedings{Lu_24_AAAI_DHVC,
  title={Deep hierarchical video compression},
  author={Lu, Ming and Duan, Zhihao and Zhu, Fengqing and Ma, Zhan},
  booktitle=AAAI,
  volume={38},
  number={8},
  pages={8859--8867},
  year={2024}
}

@article{Lu_24_arxiv_DHVC2,
  title={High-Efficiency Neural Video Compression via Hierarchical Predictive Learning},
  author={Lu, Ming and Duan, Zhihao and Cong, Wuyang and Ding, Dandan and Zhu, Fengqing and Ma, Zhan},
  journal={arXiv preprint arXiv:2410.02598},
  year={2024}
}

@inproceedings{Zhang_21_VCIP_DVCP,
  title={DVC-P: Deep video compression with perceptual optimizations},
  author={Zhang, Saiping and Mrak, Marta and Herranz, Luis and Blanch, Marc G{\'o}rriz and Wan, Shuai and Yang, Fuzheng},
  booktitle=VCIP,
  pages={1--5},
  year={2021},
  organization={IEEE}
}

@inproceedings{Li_25_ICLR_GIC,
  title={Once-for-all: Controllable generative image compression with dynamic granularity adaptation},
  author={Li, Anqi and Li, Feng and Liu, Yuxi and Cong, Runmin and Zhao, Yao and Bai, Huihui},
  booktitle=ICLR,
  year={2025}
}

@inproceedings{Zhang_25_ICCV_StableCodec,
  title={StableCodec: Taming one-step diffusion for extreme image compression},
  author={Zhang, Tianyu and Luo, Xin and Li, Li and Liu, Dong},
  booktitle = ICCV,
  year      = {2025}
}

@inproceedings{Xue_25_ICCV_DLF,
  title={DIF: Extreme image compression with dual-generative latent fusion},
  author={Xue, Naifu and Jia, Zhaoyang and Li, Jiahao and Li, Bin and Zhang, Yuan and Lu, Yan},
  booktitle = ICCV,
  year      = {2025}
}

@inproceedings{Mentzer_22_ECCV_NVCGAN,
  title={Neural video compression using gans for detail synthesis and propagation},
  author={Mentzer, Fabian and Agustsson, Eirikur and Ball{\'e}, Johannes and Minnen, David and Johnston, Nick and Toderici, George},
  booktitle=ECCV,
  pages={562--578},
  year={2022},
  organization={Springer}
}

@article{Qi_25_TCSVT_GLCVideo,
  title={Generative latent coding for ultra-low bitrate image and video compression},
  author={Qi, Linfeng and Jia, Zhaoyang and Li, Jiahao and Li, Bin and Li, Houqiang and Lu, Yan},
  journal=TCSVT,
  year={2025},
  publisher={IEEE}
}

@article{Ma_25_tomm_DiffVC,
  title={Diffusion-based perceptual neural video compression with temporal diffusion information reuse},
  author={Ma, Wenzhuo and Chen, Zhenzhong},
  journal=TOMM,
  volume={21},
  number={12},
  pages={1--22},
  year={2025},
  publisher={ACM New York, NY}
}

@inproceedings{Li_24_WCSP_EVCDiffusion,
  title={Extreme Video Compression with Prediction Using Pre-trained Diffusion Models},
  author={Li, Bohan and Liu, Yiming and Niu, Xueyan and Bait, Bo and Han, Wei and Deng, Lei and Gunduz, Deniz},
  booktitle=WCSP,
  pages={1449--1455},
  year={2024},
  organization={IEEE}
}

@article{Mao_21_TCSVT_VVCRC,
  title={High efficiency rate control for versatile video coding based on composite Cauchy distribution},
  author={Mao, Yunhao and Wang, Meng and Wang, Shiqi and Kwong, Sam},
  journal=TCSVT,
  volume={32},
  number={4},
  pages={2371--2384},
  year={2021},
  publisher={IEEE}
}

@article{Feng_24_TCSVT_RC,
  title={Content-Adaptive Rate Control Method for User-Generated Content Videos},
  author={Feng, Longtao and Yin, Qian and Ma, Siwei},
  journal=TCSVT,
  year={2024},
  publisher={IEEE}
}

@article{Liao_24_TMM_RC,
  title={Content-adaptive rate-distortion modeling for frame-level rate control in versatile video coding},
  author={Liao, Junqi and Li, Li and Liu, Dong and Li, Houqiang},
  journal=TMM,
  volume={26},
  pages={6864--6879},
  year={2024},
  publisher={IEEE}
}

@inproceedings{Zhang_24_ICLR_NRC,
  title={Neural rate control for learned video compression},
  author={Zhang, Yiwei and Lu, Guo and Chen, Yunuo and Wang, Shen and Shi, Yibo and Wang, Jing and Song, Li},
  booktitle=ICLR,
  year={2024}
}

@inproceedings{Gu_25_DCC_ARC,
  title={Adaptive rate control for deep video compression with rate-distortion prediction},
  author={Gu, Bowen and Chen, Hao and Lu, Ming and Yao, Jie and Ma, Zhan},
  booktitle=DCC,
  pages={33--42},
  year={2025},
  organization={IEEE}
}

@article{Feng_25_TCSVT_HARC,
  title={High Accuracy Rate Control for Neural Video Coding Based on Rate-Distortion Modeling},
  author={Feng, Longtao and Yin, Qian and Zhang, Jiaqi and He, Yuwen and Ma, Siwei},
  journal=TCSVT,
  year={2025},
  publisher={IEEE}
}

@inproceedings{Li_22_ICASSP_RC,
  title={Rate control for learned video compression},
  author={Li, Yanghao and Chen, Xinyao and Li, Jisheng and Wen, Jiangtao and Han, Yuxing and Liu, Shan and Xu, Xiaozhong},
  booktitle=ICASSP,
  pages={2829--2833},
  year={2022},
  organization={IEEE}
}

@inproceedings{Zhang_24_ECCV_LRC,
  title={Learned rate control for frame-level adaptive neural video compression via dynamic neural network},
  author={Zhang, Chenhao and Gao, Wei},
  booktitle=ECCV,
  pages={239--255},
  year={2024},
  organization={Springer}
}

@article{Chen_23_TCSVT_StD,
  title={Sparse-to-dense: High efficiency rate control for end-to-end scale-adaptive video coding},
  author={Chen, Jiancong and Wang, Meng and Zhang, Pingping and Wang, Shurun and Wang, Shiqi},
  journal=TCSVT,
  volume={34},
  number={5},
  pages={4027--4039},
  year={2023},
  publisher={IEEE}
}

@article{Theis_2022_arxiv_DiffC,
  author       = {Lucas Theis and
                  Tim Salimans and
                  Matthew D. Hoffman and
                  Fabian Mentzer},
  title        = {Lossy Compression with Gaussian Diffusion},
  journal      = {arXiv preprint arXiv:2206.08889},
  year         = {2022}
}

@inproceedings{Vonderfecht_25_ICLR_DiffC,
  author       = {Jeremy Vonderfecht and
                  Feng Liu},
  title        = {Lossy Compression with Pretrained Diffusion Models},
  booktitle    = ICLR,
  year         = {2025}
}

@inproceedings{Theis_22_ICML_RCC,
  author       = {Lucas Theis and
                  Noureldin Y. Ahmed},
  title        = {Algorithms for the Communication of Samples},
  booktitle    = ICML,
  volume       = {162},
  pages        = {21308--21328},
  publisher    = {},
  year         = {2022}
}

@article{Wang_25_arxiv_TGVC,
  title={T-GVC: Trajectory-Guided Generative Video Coding at Ultra-Low Bitrates},
  author={Wang, Zhitao and Man, Hengyu and Li, Wenrui and Wang, Xingtao and Fan, Xiaopeng and Zhao, Debin},
  journal={arXiv preprint arXiv:2507.07633},
  year={2025}
}

@inproceedings{Chen_25_ICCV_HyTIP,
    title     = {HyTIP: Hybrid Temporal Information Propagation for Masked Conditional Residual Video Coding},
    author    = {Chen, Yi-Hsin and Yao, Yi-Chen and Ho, Kuan-Wei and Wu, Chun-Hung and Phung, Huu-Tai and Benjak, Martin and Ostermann, Jörn and Peng, Wen-Hsiao},
    booktitle = ICCV,
    year      = {2025}
}

@inproceedings{Jiang_25_CVPR_ECVC,
  title={ECVC: Exploiting non-local correlations in multiple frames for contextual video compression},
  author={Jiang, Wei and Li, Junru and Zhang, Kai and Zhang, Li},
  booktitle=CVPR,
  pages={7331--7341},
  year={2025}
}

@inproceedings{Liao_25_MM_EHVC,
  title={EHVC: Efficient Hierarchical Reference and Quality Structure for Neural Video Coding},
  author={Liao, Junqi and Wu, Yaojun and Lin, Chaoyi and Deng, Zhipin and Li, Li and Liu, Dong and Sun, Xiaoyan},
  booktitle=ACMMM,
  year={2025}
}

@article{Black_25_arxiv_FLUX,
      title={FLUX.1 Kontext: Flow Matching for In-Context Image Generation and Editing in Latent Space},
      author={Black Forest Labs},
      journal={arXiv preprint arXiv:2506.15742},
      year={2025}
}

@inproceedings{Harsha_07_CCC_Communication,
  title={The communication complexity of correlation},
  author={Harsha, Prahladh and Jain, Rahul and McAllester, David and Radhakrishnan, Jaikumar},
  booktitle=CCC,
  pages={10--23},
  year={2007},
  organization={IEEE}
}

@inproceedings{Havasi_19_ICLR_Minimal,
  title={Minimal random code learning: Getting bits back from compressed model parameters},
  author={Havasi, Marton and Peharz, Robert and Hern{\'a}ndez-Lobato, Jos{\'e} Miguel},
  booktitle    = ICLR,
  year         = {2019}
}

@article{Bjontegaard_01_ITU_BDRate,
  title={Calculation of average PSNR differences between RD-curves},
  author={Bjontegaard, Gisle},
  journal={ITU-T SG16, Doc. VCEG-M33},
  year={2001}
}

@inproceedings{Ma_24_CVPR_DeepCache,
  title={DeepCache: Accelerating Diffusion Models for Free},
  author={Ma, Xinyin and Fang, Gongfan and Wang, Xinchao},
  booktitle=CVPR,
  year={2024}
}

@inproceedings{Gao_2025_ICCV_givic,
  title={Givic: Generative implicit video compression},
  author={Gao, Ge and Teng, Siyue and Peng, Tianhao and Zhang, Fan and Bull, David},
  booktitle=ICCV,
  pages={17356--17367},
  year={2025}
}

@inproceedings{Fang_23_NIPS_prune,
  title={Structural pruning for diffusion models},
  author={Gongfan Fang and Xinyin Ma and Xinchao Wang},
  booktitle=NeurIPS,
  year={2023},
}

@inproceedings{Ohayon_25_ICML_DDCM,
  author       = {Ohayon, Guy and Manor, Hila and Michaeli, Tomer and Elad, Michael},
  title        = {Compressed Image Generation with Denoising Diffusion Codebook Models},
  booktitle    = ICML,
  publisher    = {},
  year         = {2025}
}

@article{Sheng_25_TIP_PRQA,
  title={Prediction and reference quality adaptation for learned video compression},
  author={Sheng, Xihua and Li, Li and Liu, Dong and Li, Houqiang},
  journal=TIP,
  year={2025},
  publisher={IEEE}
}

@inproceedings{Bian_25_CVPR_SEVC,
  title={Augmented Deep Contexts for Spatially Embedded Video Coding},
  author={Bian, Yifan and Tang, Chuanbo and Li, Li and Liu, Dong},
  booktitle=CVPR,
  pages={2094--2104},
  year={2025}
}

@inproceedings{Tang_25_CVPR_DCMVC,
  title={Neural Video Compression with Context Modulation},
  author={Tang, Chuanbo and Li, Zhuoyuan and Bian, Yifan and Li, Li and Liu, Dong},
  booktitle=CVPR,
  pages={12553--12563},
  year={2025}
}

@inproceedings{Fang_25_CVPR_Tinyfusion,
  title={Tinyfusion: Diffusion transformers learned shallow},
  author={Fang, Gongfan and Li, Kunjun and Ma, Xinyin and Wang, Xinchao},
  booktitle=CVPR,
  pages={18144--18154},
  year={2025}
}

@inproceedings{Vaisman_25_arxiv_turboddcm,
  title={Turbo-DDCM: Fast and flexible zero-shot diffusion-based image compression},
  author={Vaisman, Amit and Ohayon, Guy and Manor, Hila and Elad, Miki and Michaeli, Tomer},
  booktitle=ICLR,
  volume={2026},
  pages={151475--151520},
  year={2026}
}

@article{Ma_25_arxiv_diffvc,
  title={DiffVC-OSD: One-Step Diffusion-based Perceptual Neural Video Compression Framework},
  author={Ma, Wenzhuo and Chen, Zhenzhong},
  journal={arXiv preprint arXiv:2508.07682},
  year={2025}
}

@inproceedings{Wang_03_ACSSC_MSSSIM,
  title={Multiscale structural similarity for image quality assessment},
  author={Wang, Zhou and Simoncelli, Eero P. and Bovik, Alan C.},
  booktitle=ASILOMAR,
  year={2003},
  organization={IEEE}
}

@article{Li_18_TIT_PFR,
  title={Strong functional representation lemma and applications to coding theorems},
  author={Li, Cheuk Ting and El Gamal, Abbas},
  journal=TIT,
  volume={64},
  number={11},
  pages={6967--6978},
  year={2018},
  publisher={}
}

@article{Zhou_25_VI_tvc,
  title={TVC: tokenized video compression with ultra-low bit rate},
  author={Zhou, Lebin and Ruan, Cihan and Ling, Nam and Chen, Zhenghao and Wang, Wei and Jiang, Wei},
  journal=VI,
  volume={3},
  number={1},
  pages={25},
  year={2025},
  publisher={Springer}
}

@article{Fani_25_TOMM_MAG,
  title={On Temporal Smoothness of Video Reconstruction Quality in the DCVS via Non-Uniform Sampling},
  author={Fani, Davoud and Beheshti-Shirazi, Aliasghar and Ghanbari, Mohammad and Rezaei, Esmatollah},
  journal=TOMM,
  volume={21},
  number={6},
  pages={1--26},
  year={2025},
  publisher={ACM New York, NY}
}

@article{Ma_25_TMLR_Latte,
  title={Latte: Latent Diffusion Transformer for Video Generation},
  author={Ma, Xin and Wang, Yaohui and Chen, Xinyuan and Jia, Gengyun and Liu, Ziwei and Li, Yuan-Fang and Chen, Cunjian and Qiao, Yu},
  journal=TMLR,
  year={2025}
}

@article{Zhou_24_arxiv_allegro,
  title={Allegro: Open the black box of commercial-level video generation model},
  author={Zhou, Yuan and Wang, Qiuyue and Cai, Yuxuan and Yang, Huan},
  journal={arXiv preprint arXiv:2410.15458},
  year={2024}
}

\end{document}